\documentclass{granularity_preprint}

\usepackage{amsmath,amsfonts,bm}

\def\eqref#1{equation~\ref{#1}}

\def\1{\bm{1}}

\DeclareMathAlphabet{\mathsfit}{\encodingdefault}{\sfdefault}{m}{sl}
\SetMathAlphabet{\mathsfit}{bold}{\encodingdefault}{\sfdefault}{bx}{n}

\usepackage{iftex}
\ifPDFTeX
  \usepackage[utf8]{inputenc} 
\fi
\usepackage[T1]{fontenc}    
\usepackage{hyperref}       
\usepackage{url}            
\usepackage{booktabs}       
\usepackage{amsfonts}       
\usepackage{nicefrac}       
\usepackage[nopatch=footnote]{microtype}
\usepackage{xcolor}         
\hypersetup{
  colorlinks=true,
  linkcolor=paperBlue,
  citecolor=paperBlue,
  urlcolor=paperBlue,
  filecolor=paperBlue,
}
\usepackage{amsmath}
\usepackage{tabularx}
\usepackage{array}
\usepackage{longtable}
\usepackage{colortbl}       
\usepackage{multirow}       
\usepackage{setspace}       
\usepackage[ruled,linesnumbered,norelsize]{algorithm2e}

\definecolor{microtint}{RGB}{232,240,250}   
\definecolor{neutraltint}{RGB}{245,245,245} 
\definecolor{macrotint}{RGB}{252,238,228}   
\definecolor{headertint}{RGB}{222,228,238}  
\definecolor{microaccent}{RGB}{40,90,160}   
\definecolor{macroaccent}{RGB}{180,80,30}   

\colorlet{takeawaybg}{paperTint}

\usepackage{tcolorbox}
\tcbuselibrary{breakable, skins}

\tcbset{
  aibox/.style={
    width=\linewidth,
    top=1pt, bottom=1pt, left=5pt, right=5pt,
    before skip=2pt, after skip=2pt,
    colback=takeawaybg,
    colframe=paperBlue!40,
    boxrule=0.5pt,
    arc=2pt,
    fontupper=\small\itshape,
    breakable,
  }
}
\newtcolorbox{AIbox}{aibox}

\colorlet{algkw}{paperBlue}
\colorlet{algcm}{paperGray}
\colorlet{algcap}{paperBlue}

\SetKwSty{algkwfont}
\SetFuncSty{algkwfont}
\SetCommentSty{algcmfont}
\SetAlCapFnt{\color{algcap}\bfseries}
\SetAlCapNameFnt{\color{algcap}\bfseries}
\IncMargin{1.5em}
\newtcolorbox{promptbox}[1]{
    colback=paperTint,         
    colframe=paperBlue,       
    coltitle=white,            
    title=\textbf{#1},         
    fontupper=\ttfamily\small, 
    halign=flush left,        
    verbatim,                  
    boxrule=0.5mm,
    arc=2mm,
    top=2mm, bottom=2mm, left=2mm, right=2mm
}

\title{\paperaccent{The Granularity Axis}: A Micro-to-Macro \mbox{Latent} Direction for Social Roles in \mbox{Language} Models}

\author{%
\textbf{Chonghan Qin}$^{1}$,
\textbf{Xiachong Feng}$^{1}$,
\textbf{Ziyun Song}$^{2}$,
\textbf{Xiaocheng Feng}$^{2}$,
\textbf{Jing Xiong}$^{1}$,
\textbf{Lingpeng Kong}$^{1}$ \\
$^{1}$The University of Hong Kong \quad
$^{2}$Harbin Institute of Technology
}

\correspondence{\email{qinch@connect.hku.hk}, \email{fxc@hku.hk}}
\checkdata[Code]{\url{https://github.com/chonghanqin/Granularity-Axis/}}

\begin{document}

\maketitle
\setcounter{footnote}{0}

\begin{abstract}
    Large language models (LLMs) are prompted to adopt social roles from individuals to institutions, but whether their representations encode this difference in social \emph{granularity} remains unclear. We discover a contrast-based \emph{Granularity Axis}, the difference between mean macro- and micro-role hidden states, that aligns with the dominant geometry of role space: in Qwen3-8B, its cosine with PC1 is $0.972$, and PC1 explains $52.6\%$ of the variance. Across 75 roles at five levels, with 90,000 role-conditioned and 1,200 default responses per model, projections increase monotonically and remain stable across layers, prompts, and response-quality filters. The structure transfers to Llama-3.1-8B-Instruct and scales to Qwen3-32B and Llama-3.1-70B-Instruct, whose PC1 shares are $45.99\%$ and $31.23\%$. Independent five-rater validation ($\alpha=0.961$) preserves the ordering; identical continuations and affect/formality controls confirm scale-related variation beyond generated wording. The axis is also \emph{behaviorally causal}: positive steering moves Llama from $2.00$ to $3.17$ on a five-point macro scale for prompts admitting local responses. Steering strength and stability depend on the model's default operating regime. Together, these findings establish social granularity as a representational primitive: a continuous, ordered, causally manipulable direction organizing role-conditioned generation across model families.
\end{abstract}

\section{Introduction}

Large language models (LLMs) adapt their behavior under prompting \citep{gpt5, llama3, qwen3, gemini3, claude}, motivating their use in social simulation \citep{generative_agents_human, social_sim_survey, llm_social_sim}. Yet representational validity, survey bias, and overly rationalized behavior remain concerns \citep{Santurkar2023WhoseOD, Bisbee2024SyntheticRF, social_desirability_bias, more_rational_than_humans}. These concerns depend on what a model internally represents when prompted to \emph{be} someone: stylistic mimicry and a distinct perspective have different implications for simulation.

Role conditioning can elicit the voices of a worried parent, community organizer, hospital administrator, or central bank governor \citep{shao2023characterllm, wang2024rolellm, JiangPersonaLLMIT, Wang2023InCharacterEP, Tu2024CharacterEvalAC, Kong2023BetterZR}. We ask: \emph{does an LLM internally distinguish the granularity of these social roles, or realize them through a shared role-playing template?}

Social roles differ in agency, time horizon, and structural constraints \citep{coleman1990foundations, Schelling1978MicromotivesAM, Granovetter1978ThresholdMO}. Micro roles emphasize personal experience and immediate concerns; macro roles involve coordination, institutional constraints, and long-term strategy. We call a mismatch between the scale a context requires and the scale a model adopts \emph{granularity confusion}: individual reasoning where systemic analysis is needed, or institutional abstraction where local judgment is appropriate. If a central bank governor and worried parent reason at the same scale, a nominally multi-perspective simulation can conceal representational collapse despite different wording.

\begin{figure*}[t]
    \centering
    \includegraphics[width=\textwidth]{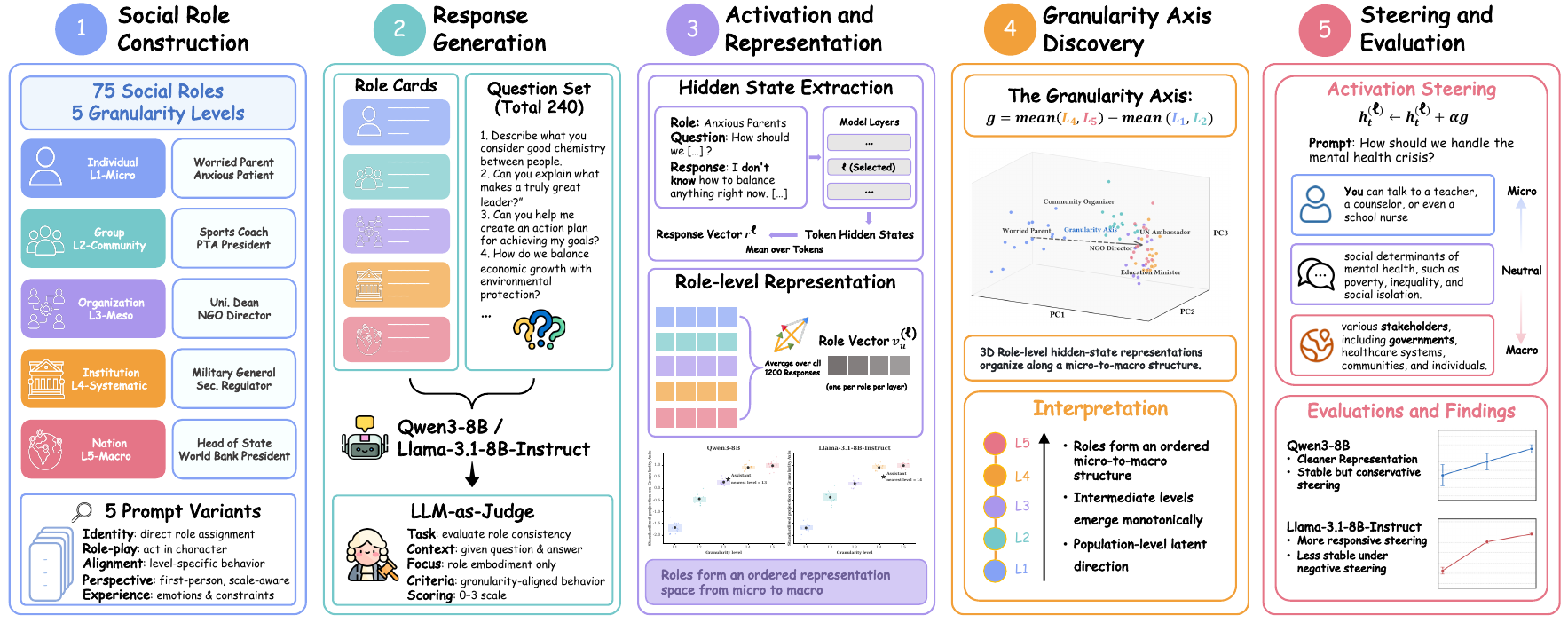}
    \caption{\textbf{Overview of the Granularity Axis pipeline.}
    We construct ordered social roles, collect role-conditioned responses, extract role-level hidden-state representations, define a contrast-based Granularity Axis, and test its behavioral effect through activation steering.}
    \label{fig:method}
\end{figure*}

Interpretability research motivates looking for this distinction in low-dimensional activation structure \citep{park2024linear, Park2024TheGO, Sakata2026LinearRO}. The \emph{Assistant Axis} tracks deviation from the default assistant persona \citep{assistant_axis}, while activation steering establishes that latent directions can causally shape behavior \citep{steering, repe, caa}. This predicts that social granularity should surface as a single, ordered direction in activation space.

We construct 75 roles spanning five granularity levels, collect responses to shared questions under multiple prompts, and average their hidden states into role vectors. The \emph{Granularity Axis} contrasts mean macro- and micro-role vectors (Figure~\ref{fig:method}). Three findings establish its significance. First, it captures the \emph{dominant geometric axis} of role space: Qwen3-8B axis--PC1 cosine is $0.972$, PC1 explains $52.6\%$ of variance, and projections increase across all five levels. Second, the ordering survives layer, endpoint, prompt, held-out, and quality-filter variations, transfers to Llama-3.1-8B, and persists in Qwen3-32B and Llama-3.1-70B. Third, the axis is \emph{behaviorally causal}: activation interventions shift output granularity in the predicted direction, with model-dependent strength and stability.

\begin{figure*}[t]
    \centering
    \includegraphics[width=0.95\textwidth]{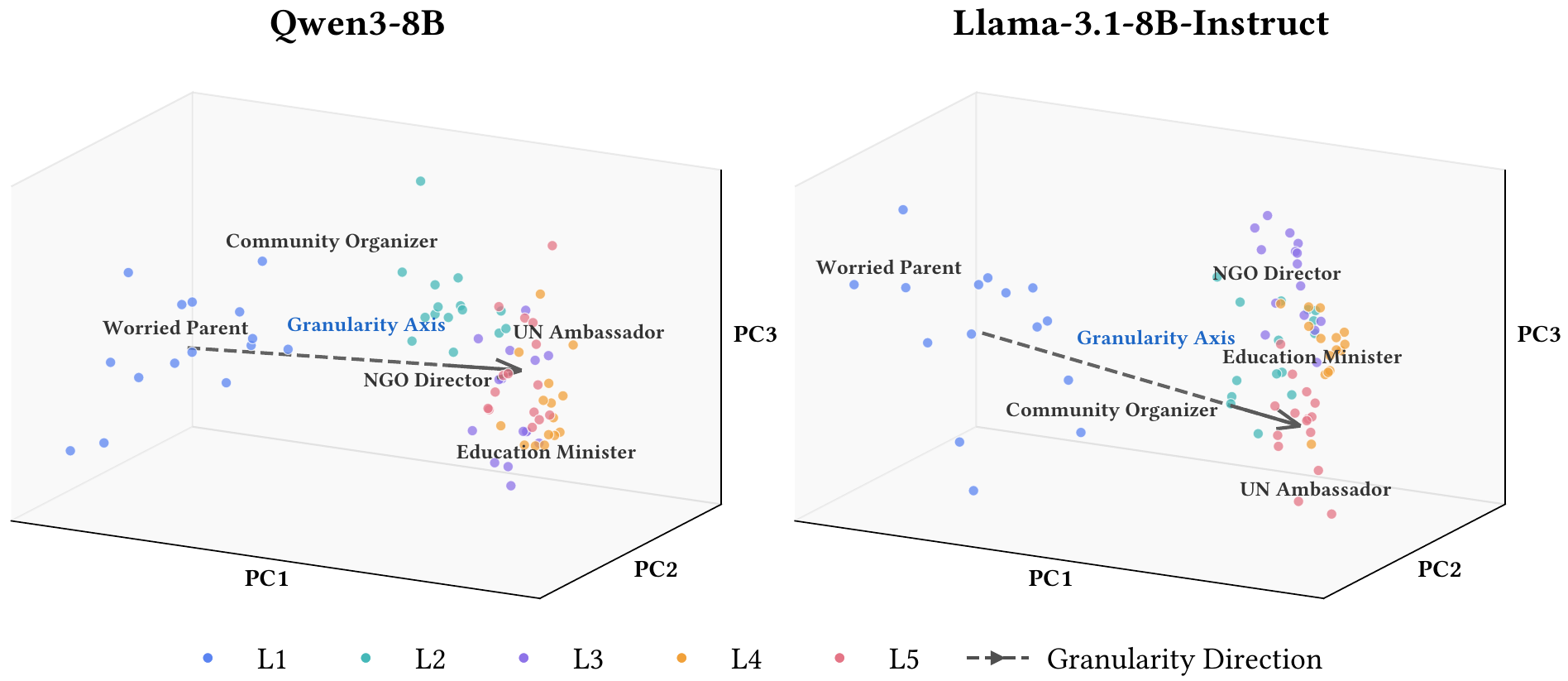}
    \caption{\textbf{Role representation space.}
    Role-level hidden-state representations organize along a micro-to-macro structure. Colors indicate granularity level (L1--L5), and the dashed arrow denotes the contrast-defined Granularity Axis from micro-level to macro-level roles.}
    \label{fig:role_representation}
\end{figure*}

Independent five-rater validation, identical teacher-forced continuations, and matched affect/formality controls further establish the ordering (\S\ref{subsec:scale-validation}--\ref{subsec:construct-results}). These findings identify social granularity as a \emph{representational primitive}, beyond a stylistic surface variable: role conditioning operates over a continuous social-scale manifold, and intervening on that direction makes scale tunable. They motivate auditing multi-agent simulations for granularity confusion, controlling social scale in personal-support or policy dialogues, and extending contrast-and-project to formality, time horizon, and risk aversion.

\section{The Granularity Axis}
\label{sec:method}

We construct a contrast-based direction and validate its geometry and causal effects (Algorithm~\ref{alg:granularity-axis}).

\subsection{Problem Setting}
\label{subsec:problem-setting}
For language model $\mathcal{M}$ with hidden dimension $d$, let $u$ be a role with level $y(u)\in\{1,\dots,5\}$ (micro to macro), $s$ a role prompt, $q$ a shared question, and $r$ a generated response. We seek a direction in the activations induced by $(u,s,q)$ that is (i) representationally meaningful, (ii) aligned with the dominant geometry of role space, and (iii) causally relevant under steering.

\begin{algorithm}[t]
\setstretch{0.92}
\caption{Granularity Axis: construction, validation, and steering.}
\label{alg:granularity-axis}
\KwIn{Language model $\mathcal{M}$; ordered role set $\mathcal{U}$ with levels $y(u)\!\in\!\{1,\dots,5\}$; prompt variants $\mathcal{S}$; shared question set $\mathcal{Q}$; analysis layer $\ell$; intervention layer $\ell^\ast$; steering strength $\alpha$.}
\KwOut{Granularity Axis $g^{(\ell)}$; role projections $\{p_u\}$; PC1 alignment $\cos(g^{(\ell)},w_1)$; steered generations.}
\tcp{Stage 1: response collection (\S\ref{subsec:roles})}
\lFor{$(u,s,q)\in\mathcal{U}\times\mathcal{S}\times\mathcal{Q}$}{generate $r_{u,s,q}\sim\mathcal{M}(\cdot\mid u,s,q)$}
\tcp{Stage 2: role-level representations (\S\ref{subsec:axis})}
\lFor{each response $(u,s,q)$}{$v^{(\ell)}_{u,s,q}\leftarrow \tfrac{1}{T}\sum_{t=1}^{T} h^{(\ell)}_t$}
\lFor{each role $u$}{$v^{(\ell)}_u \leftarrow \tfrac{1}{|\mathcal{R}(u)|}\sum_{(s,q)\in\mathcal{R}(u)} v^{(\ell)}_{u,s,q}$}
\tcp{Stage 3: contrast-based axis (\S\ref{subsec:axis})}
$g^{(\ell)} \leftarrow \tfrac{1}{|\mathcal{U}_{\text{macro}}|}\!\sum_{u\in\mathcal{U}_{\text{macro}}}\! v^{(\ell)}_u \;-\; \tfrac{1}{|\mathcal{U}_{\text{micro}}|}\!\sum_{u\in\mathcal{U}_{\text{micro}}}\! v^{(\ell)}_u$\;
\tcp{Stage 4: geometric validation (\S\ref{subsec:axis})}
form $V^{(\ell)}\!=\![v^{(\ell)}_u]_{u\in\mathcal{U}}$, center, apply PCA $\tilde V^{(\ell)} = U\Sigma W^{\!\top}$; report $\cos(g^{(\ell)},w_1)$ and $p_u = \langle v^{(\ell)}_u,\, g^{(\ell)}/\|g^{(\ell)}\|\rangle$\;
\tcp{Stage 5: causal probe via activation steering (\S\ref{subsec:steering})}
\lFor{generated token $t$}{$\hat h^{(\ell^\ast)}_t \leftarrow h^{(\ell^\ast)}_t + \alpha\, g^{(\ell^\ast)}$}
\end{algorithm}

\subsection{Ordered Social Roles and Response Collection}
\label{subsec:roles}
Our 75 roles span five levels, with 15 per level: \emph{Individual}, \emph{Group/Community}, \emph{Organization}, \emph{Institution}, and \emph{Nation/Super-Actor}. Examples are Worried Parent, Community Organizer, Hospital Administrator, Central Bank Governor, and World Bank President (Table~\ref{tab:role-taxonomy}). Levels are ordinal, non-equidistant, and defined by represented entity, constituency, and routine decision scope rather than emotion, prestige, or style. The inventory was LLM-assisted and reviewed by one author; independent five-rater validation is reported separately (Appendix~\ref{app:role-taxonomy}).

Five prompt variants change instruction style while preserving role identity: direct assignment, role-play, worldview/priority, first-person scale/time-horizon, and authenticity/constraints (Figure~\ref{pmt:role-responses}). Combining each with 240 shared questions from \citet{assistant_axis} yields $75\times5\times240=90{,}000$ role-conditioned responses and $5\times240=1{,}200$ default responses per model, with $r\sim\mathcal{M}(\cdot\mid u,s,q)$. Role-adherence scores on a 0--3 scale support quality-filter ablations (Figure~\ref{pmt:role-eval}).

\subsection{Role Representations and the Granularity Axis}
\label{subsec:axis}
At each layer $\ell$, we mean-pool assistant-token states, $v_{u,s,q}^{(\ell)}=\tfrac{1}{T}\sum_{t=1}^{T}h_t^{(\ell)}$, and average the response set $\mathcal{R}(u)$ into $v_u^{(\ell)}=\tfrac{1}{|\mathcal{R}(u)|}\sum_{(s,q)\in\mathcal{R}(u)}v_{u,s,q}^{(\ell)}$. Following the Assistant Axis contrast construction, let $\mathcal{U}_{\rm micro}=\{u:y(u)\in\{1,2\}\}$ and $\mathcal{U}_{\rm macro}=\{u:y(u)\in\{4,5\}\}$. Their mean difference defines $g^{(\ell)}=\tfrac{1}{|\mathcal{U}_{\rm macro}|}\sum_{u\in\mathcal{U}_{\rm macro}}v_u^{(\ell)}-\tfrac{1}{|\mathcal{U}_{\rm micro}|}\sum_{u\in\mathcal{U}_{\rm micro}}v_u^{(\ell)}$. Only L3 is wholly excluded from this contrast. We test whether projections rise monotonically across all five levels.

We center the 75-role matrix $V^{(\ell)}\in\mathbb{R}^{75\times d}$ and compute PCA, $\tilde V^{(\ell)}=U\Sigma W^\top$, testing axis--PC1 cosine and correlation of projections $p_u=\langle v_u^{(\ell)},g^{(\ell)}/\|g^{(\ell)}\|\rangle$ with $y(u)$. The default is excluded from PCA and contrast construction. PC1 share describes the role matrix; it differs from variance explained by the contrast itself. Alternative endpoints, the Assistant Axis, and random directions provide controls.

Monotonic means test group order; role-level Spearman measures agreement across individual roles. Axis--PC1 cosine tests whether the labeled contrast recovers the dominant unlabeled direction, while PC1 share measures variance concentration. For scaling, the participation ratio $(\sum_k\lambda_k)^2/\sum_k\lambda_k^2$ of PCA eigenvalues $\lambda_k$ increases as variance spreads across components. These metrics distinguish a persistent scale direction from an entirely one-dimensional role space.

\subsection{Activation Steering}
\label{subsec:steering}
At intervention layer $\ell^\ast$, selected by a layer sweep, we add $\alpha g^{(\ell^\ast)}$ to each generated-token activation: $\hat h_t^{(\ell^\ast)}=h_t^{(\ell^\ast)}+\alpha g^{(\ell^\ast)}$. Positive $\alpha$ pushes toward institutional/systemic reasoning; negative $\alpha$ toward individual/local reasoning. Prompt encoding is unchanged. We measure direction, strength, symmetry, and stability at fixed prompts.

\subsection{Controlled Forward-pass Validation}
\label{subsec:construct-method}
In both 8B models, we extract final role-prompt, final question, pre-first-assistant, and pooled identical teacher-forced continuation states. Continuation tokens and lengths are fixed across roles, without free generation or role-specific truncation. Eight domain chains cross five scope levels with affect/formality conditions and 24 questions. Axes trained on seven neutral domains are evaluated on the eighth. Regression includes level, affect, formality, level-by-affect/formality interactions, and domain effects; cluster bootstrap estimates uncertainty and joint scale partial $R^2$ (Appendix~\ref{app:construct-validity}).

The original-role test uses 48 deterministic questions and five prompts: 18,000 inputs per model. Cached default responses supply at most 256 identical continuation tokens. Contrasts are rebuilt at each position, testing scale encoding before generation and during identical-token processing without assuming one shared direction. Qwen's role-prompt axis, for example, has cosine $0.160$ with its response-mean axis despite strong ordering. The matched-domain experiment separately tests transfer beyond the domains used to construct its axis.

\section{Experiments}
\label{sec:experiments}

\begin{figure*}[t]
    \centering
    \includegraphics[width=0.95\textwidth]{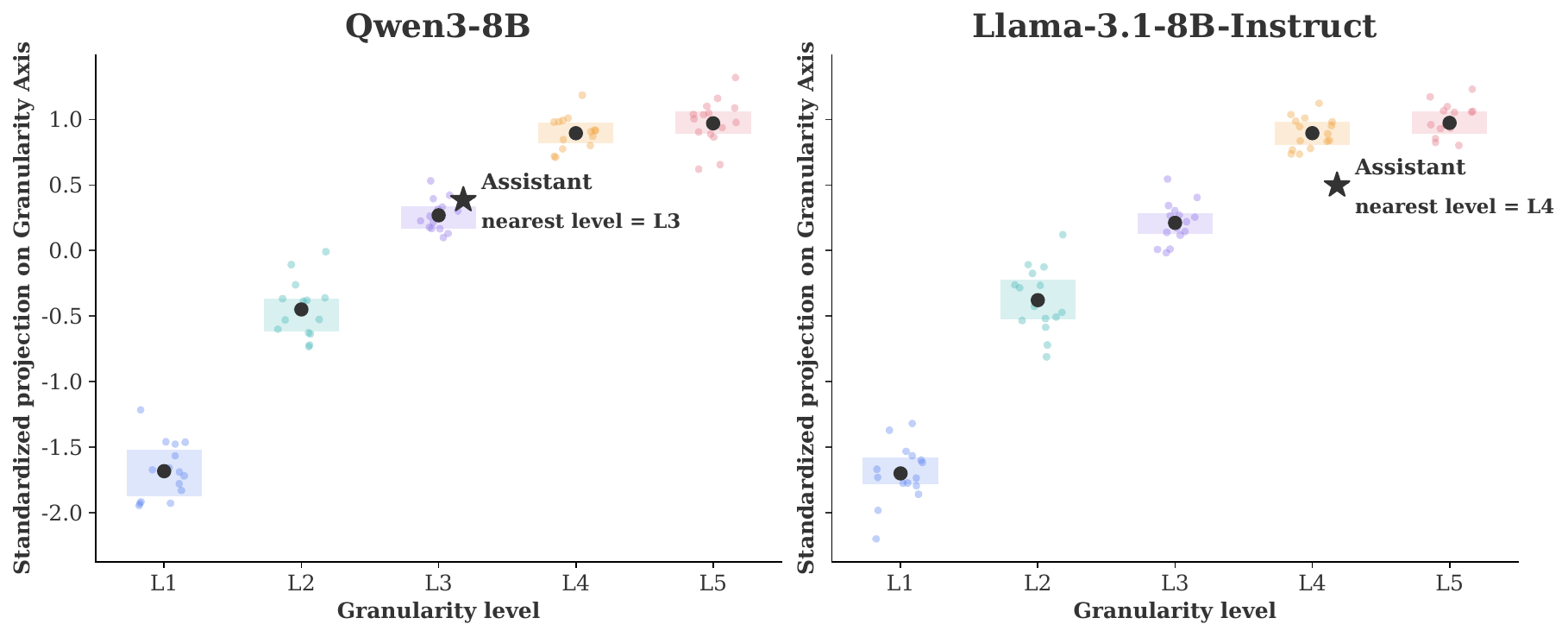}
    \caption{\textbf{Ordered projections on the Granularity Axis.}
    Points are roles grouped by granularity level; black circles mark level means, shaded bands within-level spread, stars the default assistant. Projections rise monotonically L1$\to$L5 in both models; the default sits in a meso-to-macro region (near L3 in Qwen3-8B, L4 in Llama-3.1-8B-Instruct).}
    \label{fig:ordered_projections}
\end{figure*}

\subsection{Experimental Setup}
\label{subsec:experiment-setup}
We apply the same pipeline to Qwen3-8B and Llama-3.1-8B-Instruct, then Qwen3-32B and Llama-3.1-70B-Instruct (\S\ref{subsec:scale-validation}). Each model has 1,200 responses per role/default condition (\S\ref{subsec:roles}). Layer~18 lies in the stable middle-layer regime of both 8B models.

Steering uses Layer~18, coefficients $\{-4,0,+4\}$, and greedy decoding on 40 \emph{generic} social-policy prompts and 12 \emph{micro-targeted} prompts admitting local responses (Appendix~\ref{app:steering-prompt-sets}). The latter expose effects masked by Qwen's macro-oriented generic baseline. The primary judge is \texttt{gpt-5.4-mini} \citep{gpt54mini}, checked against \texttt{gemini-3.1-flash-lite-preview} \citep{gemini31flashlite}; Figure~\ref{pmt:steering-eval} gives the rubric. Appendix~\ref{app:reproduce} records compute, questions, and licenses.

\subsection{Representation Results}
\begin{table*}[t]
\centering
\scriptsize
\setlength{\tabcolsep}{3pt}
\renewcommand{\arraystretch}{1.1}
\arrayrulecolor{black!60}
\caption{\textbf{Main representation results at the target layer.}
In both models, the Granularity Axis aligns with PC1 and strongly correlates with the five-level ordering; the default assistant sits in a meso-to-macro region. Asst.\ Level cells reuse the L1$\to$L5 tints of Table~\ref{tab:level_means}.}
\label{tab:main_representation}
\begin{tabular*}{\textwidth}{
@{\extracolsep{\fill}}
l c c c c c c
@{}}
\toprule
\rowcolor{headertint}
\textbf{Model} & \textbf{Layer} & \textbf{PC1 Var.} & $\boldsymbol{\cos(g,\mathrm{PC1})}$ & \textbf{Spearman} & \textbf{Pearson} & \textbf{Asst.\ Level} \\
\midrule
\textbf{Qwen3-8B}              & $18$ & $52.57\%$ & $0.9720$ & $0.9472$ & $0.9414$ & \cellcolor{microtint!50!macrotint}L3 \\
\textbf{Llama-3.1-8B-Instruct} & $18$ & $42.46\%$ & $0.9596$ & $0.9459$ & $0.9373$ & \cellcolor{microtint!25!macrotint}L4 \\
\bottomrule
\end{tabular*}
\arrayrulecolor{black}
\end{table*}

Figures~\ref{fig:role_representation}--\ref{fig:ordered_projections} and Table~\ref{tab:main_representation} show a dominant micro-to-macro organization. At Layer~18, Qwen/Llama axis--PC1 cosines are $0.9720/0.9596$, PC1 shares are $52.57\%/42.46\%$, and Spearman/Pearson exceed $0.93$. Mean projections increase through L1--L5, with saturation at L4--L5 (Table~\ref{tab:level_means}). Qwen exhibits the more concentrated geometry. Default assistants lie near L3 (Qwen) and L4 (Llama), helping explain steering asymmetry (Appendix~\ref{app:default-assistant-placement}). These results satisfy criteria (i)--(ii).

\begin{table}[t]
\centering
\footnotesize
\setlength{\tabcolsep}{6pt}
\renewcommand{\arraystretch}{1.25}
\arrayrulecolor{black!60}
\caption{\textbf{Mean projections onto the Granularity Axis per granularity level.} Both models exhibit monotonic L1$\to$L5 ordering with saturation between L4 and L5, despite very different absolute scales. Column tints follow the micro$\to$macro spectrum used in Tables~\ref{tab:steering_results} and~\ref{tab:qual_example}.}
\label{tab:level_means}
\begin{tabular*}{\linewidth}{
@{\extracolsep{\fill}}
l
>{\columncolor{microtint}}c
>{\columncolor{microtint!75!macrotint}}c
>{\columncolor{microtint!50!macrotint}}c
>{\columncolor{microtint!25!macrotint}}c
>{\columncolor{macrotint}}c
@{}}
\toprule
\rowcolor{headertint}
\textbf{Model} & \textbf{L1} & \textbf{L2} & \textbf{L3} & \textbf{L4} & \textbf{L5} \\
\midrule
\textbf{Qwen3-8B}              & $0.5425$  & $11.1058$ & $17.2525$ & $22.6148$ & $23.2603$ \\
\textbf{Llama-3.1-8B-Instruct} & $-0.8784$ & $1.2059$  & $2.1362$  & $3.2149$  & $3.3386$  \\
\bottomrule
\end{tabular*}
\arrayrulecolor{black}
\end{table}

\subsection{Larger Models and Independent Taxonomy Validation}
\label{subsec:scale-validation}
\begin{table}[t]
\centering
\small
\setlength{\tabcolsep}{4pt}
\renewcommand{\arraystretch}{1.15}
\caption{\textbf{Ordering persists while role geometry becomes more distributed.} All four instruction-tuned checkpoints have strictly increasing level means under original and median-consensus labels. PC1 share and effective rank use all 75 role vectors and are label-invariant. Large-model layers were fixed by relative depth before their analyses; Llama names abbreviate the Instruct checkpoints.}
\label{tab:model-scale}
\begin{tabular*}{\linewidth}{@{\extracolsep{\fill}}lrrrrrr@{}}
\toprule
\textbf{Model} & \textbf{Layer} & \textbf{PC1 \%} & \textbf{Rank} & \multicolumn{2}{c}{\textbf{Original labels}} & \textbf{Consensus} \\
 & & & & $|\cos(g,w_1)|$ & $\rho$ & $|\cos|$ / $\rho$ \\
\midrule
Qwen3-8B & 18 & 52.57 & 3.39 & .9720 & .9472 & .9830 / .9358 \\
Qwen3-32B & 32 & 45.99 & 4.42 & .9814 & .9472 & .9893 / .9384 \\
Llama-3.1-8B & 18 & 42.46 & 5.03 & .9596 & .9459 & .9735 / .9415 \\
Llama-3.1-70B & 45 & 31.23 & 8.56 & .9633 & .9655 & .9777 / .9639 \\
\bottomrule
\end{tabular*}
\end{table}

The scaling comparison retains the same roles, prompts, and questions, with equal response weights and no quality filter. Primary layers are selected by relative depth: Qwen Layer~18 of 36 maps to Layer~32 of 64, and Llama Layer~18 of 32 maps to Layer~45 of 80. Target layers are therefore fixed before inspecting the larger-model results.

At pre-specified relative depths (Qwen3-32B Layer~32; Llama-70B Layer~45), ordering and axis--PC1 alignment persist (Table~\ref{tab:model-scale}). PC1 share falls from $52.57\%$ to $45.99\%$ in Qwen and $42.46\%$ to $31.23\%$ in Llama, while effective rank rises from $3.39$ to $4.42$ and $5.03$ to $8.56$. Ordering is monotonic at $59/64$ and $80/80$ layers, respectively: the axis persists as role geometry becomes more distributed. Both larger defaults lie nearest L3 (Appendix~\ref{app:scaling}).

Five independent blinded raters yield ordinal Krippendorff's $\alpha=0.961$ (95\% CI $[0.933,0.979]$), pairwise quadratic-weighted $\kappa=0.964$, and $73.3\%$ unanimous agreement. Median labels equal strict-majority labels for every role and agree with original labels on $88\%$, without author-label tie-breaking. All four models remain monotonic under consensus labels (Table~\ref{tab:model-scale}); all 26,244 pre-specified boundary assignments preserve ordering in both 8B models (Appendix~\ref{app:role-taxonomy}).

Raters saw descriptions without original labels or model results and were not asked to balance levels. Nine roles move upward by one level, yielding consensus counts $15/11/18/12/19$. Recomputing contrasts from these labels, with PCA over the unchanged role matrix, tests whether ordering depends on the original assignments. The separate boundary analysis tests specified reassignments rather than independent human judgments.

\subsection{Construct Validity and Additional Geometric Controls}
\label{subsec:construct-results}
With identical continuations, Qwen/Llama Spearman remains $0.934/0.964$: generated-answer lexical variation is unnecessary for ordering. Matched-domain continuation scale coefficients are $0.979$ [$0.898,1.064$] and $0.253$ [$0.238,0.267$], with joint scale partial $R^2=0.036/0.153$ after affect/formality controls. Across all eight model--position comparisons, held-out-domain mean Spearman spans $0.887$--$1.000$ (Appendix~\ref{app:construct-validity}).

The original-role test fixes continuation content, whereas the matched-domain factorial crosses scope with affect and formality. Each fit contains 3,840 observations. Partial $R^2$ jointly tests level and its two attribute interactions; confidence intervals resample domain--question clusters, retaining all levels and conditions. Thus the continuation test and factorial regression provide complementary evidence rather than two estimates of the same quantity.

Joint partial $R^2$ falls from $0.937/0.743$ at the role-prompt position to $0.036/0.153$ during continuations, while continuation held-out-domain mean Spearman remains $0.925/1.000$. Strong ordinal transfer thus coexists with lower scale-related variance at some positions: ordering and explained variation measure different properties of the representation.

Granularity/Assistant cosine is $0.2253/0.3492$; removing Assistant preserves ordering ($\rho=0.9481/0.9533$). Removing Granularity leaves residual PC1 with $12.80\%/12.06\%$ of original variance. Permutation tests support axis--PC1 alignment, while rebuilt-contrast rank and L4--L5 separation tests are nonsignificant; Appendix~\ref{app:additional-validation} reports all controls and residual structure.

After Assistant removal, residual Granularity--PC1 cosine remains $0.9399/0.8955$, whereas Assistant after Granularity removal is nonmonotonic. This establishes ordering beyond the Assistant direction without requiring orthogonality. Fixed-projection permutation tests condition on the observed axis; rebuilt-contrast nulls also account for endpoint selection, explaining why their rank evidence is weaker.

\begin{table*}[t]
\centering
\scriptsize
\setlength{\tabcolsep}{4pt}
\renewcommand{\arraystretch}{1.3}
\arrayrulecolor{black!60}
\caption{\textbf{Steering shifts judged output granularity in a model- and prompt-dependent way.}
Mean $\pm$ SEM \texttt{granularity\_overall} from \texttt{gpt-5.4-mini} (1--5; higher = more macro), with SEM over prompts. $\Delta_{\pm 4}$ are deltas from the unsteered baseline; Deg. reports degeneration rates at $\alpha\!=\!-4,0,+4$ (filtered scores in Appendix~\ref{app:additional-steering}). Tints follow the micro-to-macro spectrum of Table~\ref{tab:qual_example}.}
\label{tab:steering_results}
\begin{tabular*}{\textwidth}{
@{\extracolsep{\fill}}
ll
>{\columncolor{microtint}}c
>{\columncolor{neutraltint}}c
>{\columncolor{macrotint}}c
cc
c
@{}}
\toprule
\rowcolor{headertint}
\textbf{Model} & \textbf{Prompt set} &
\textcolor{microaccent}{$\boldsymbol{\alpha\!=\!-4}$} &
$\boldsymbol{\alpha\!=\!0}$ &
\textcolor{macroaccent}{$\boldsymbol{\alpha\!=\!+4}$} &
\textcolor{microaccent}{$\boldsymbol{\Delta_{-4}}$} &
\textcolor{macroaccent}{$\boldsymbol{\Delta_{+4}}$} &
\textbf{Deg. $(-/0/+)$} \\
\midrule
\multirow{2}{*}{\textbf{Qwen3-8B}}
 & Generic        & $4.821{\pm}.062$ & $4.900{\pm}.048$ & $4.975{\pm}.025$ & \textcolor{microaccent}{$-0.080$} & \textcolor{macroaccent}{$+0.075$} & .000/.000/.000 \\
 & Micro-targeted & $2.000{\pm}.000$ & $2.167{\pm}.112$ & $2.667{\pm}.256$ & \textcolor{microaccent}{$-0.167$} & \textcolor{macroaccent}{$+0.500$} & .000/.000/.000 \\
\addlinespace[2pt]
\cmidrule(lr){1-8}
\addlinespace[2pt]
\multirow{2}{*}{\textbf{Llama-3.1-8B-Instruct}}
 & Generic        & $3.125{\pm}.161$ & $4.525{\pm}.088$ & $4.900{\pm}.048$ & \textcolor{microaccent}{$-1.400$} & \textcolor{macroaccent}{$+0.375$} & .425/.000/.025 \\
 & Micro-targeted & $1.583{\pm}.149$ & $2.000{\pm}.000$ & $3.167{\pm}.241$ & \textcolor{microaccent}{$-0.417$} & \textcolor{macroaccent}{$+1.167$} & .000/.083/.000 \\
\bottomrule
\end{tabular*}
\arrayrulecolor{black}
\end{table*}

\subsection{Steering Results}
\label{subsec:steering-results}
\begin{table*}[ht]
\centering
\scriptsize
\setlength{\tabcolsep}{4pt}
\renewcommand{\arraystretch}{1.2}
\renewcommand{\tabularxcolumn}[1]{m{#1}} 
\arrayrulecolor{black!60}
\caption{Manually selected illustrative responses under no steering and granularity steering; these cases are not used for quantitative conclusions. Generic-setting degeneration rates for $\alpha=-4/0/+4$ are $0/0/0$ (Qwen) and $0.425/0/0.025$ (Llama). Excerpts are quoted verbatim from the model outputs, with omissions indicated by `[...]'. Boldface marks the most salient phrases revealing granularity, with \textcolor{microaccent}{blue} for micro-leaning content and \textcolor{macroaccent}{orange} for macro-leaning content. Llama-3.1-8B-Instruct shows a clearer micro-to-macro shift, whereas Qwen3-8B remains comparatively macro-oriented across all three conditions.}
\label{tab:qual_example}
\begin{tabularx}{\textwidth}{
@{}
>{\centering\arraybackslash}m{0.018\textwidth}
>{\columncolor{microtint}\raggedright\arraybackslash}X
>{\columncolor{neutraltint}\raggedright\arraybackslash}X
>{\columncolor{macrotint}\raggedright\arraybackslash}X
@{}}
\toprule
\rowcolor{headertint}
&
\textcolor{microaccent}{$\boldsymbol{\leftarrow}$~\textbf{Micro}~($\alpha\!=\!-4$)} &
\textbf{Base}~($\alpha\!=\!0$) &
\textcolor{macroaccent}{\textbf{Macro}~($\alpha\!=\!+4$)~$\boldsymbol{\rightarrow}$} \\
\midrule
\rowcolor{white}
\multicolumn{4}{@{}l@{}}{\textit{Prompt:}~\textbf{\textit{How should we handle the mental health crisis?}}} \\
\cmidrule(lr){1-4}
\rotatebox[origin=c]{90}{\textbf{Llama-3.1-8B-Instruct}}
&
Get help when you need it: If you're struggling with your mental health, don't be afraid to \textbf{\textcolor{microaccent}{reach out to a friend, family member, or a professional for help}}. [...] Get help at school: [...] \textbf{\textcolor{microaccent}{You can talk to a teacher, a counselor, or even a school nurse}}. [...] Take care of your family.
&
Handling the mental health crisis requires a multi-faceted approach that involves \textbf{individuals, communities, organizations, and governments working together}. [...] Address systemic issues. [...] \textbf{social determinants of mental health, such as poverty, inequality, and social isolation}. [...] Invest in research and innovation.
&
Addressing the mental health crisis requires a comprehensive and multi-faceted approach that involves \textbf{\textcolor{macroaccent}{various stakeholders, including governments, healthcare systems, communities, and individuals}}. [...] \textbf{\textcolor{macroaccent}{Public education campaigns, media outreach, and social media initiatives}}. [...] \textbf{\textcolor{macroaccent}{Mental health parity laws and regulations}}. \\
\addlinespace[3pt]
\cmidrule(lr){1-4}
\addlinespace[2pt]
\rotatebox[origin=c]{90}{\textbf{Qwen3-8B}}
&
Handling the mental health crisis requires a comprehensive, multi-faceted approach that involves individuals, communities, healthcare systems, and policymakers. [...] \textbf{\textcolor{microaccent}{Build Support Networks}}: Encourage people to maintain strong social connections and seek help when needed. [...] \textbf{\textcolor{microaccent}{Support Families and Caregivers}}: [...] Offer resources to help families understand and support their loved ones.
&
Handling the mental health crisis requires a comprehensive, multi-faceted approach that addresses both \textbf{individual and systemic challenges}. [...] \textbf{Governments and organizations should invest in mental health infrastructure}. [...] \textbf{Social Determinants of Health}: Tackle issues like poverty, discrimination, housing instability, and trauma.
&
Handling the mental health crisis requires a comprehensive, multi-faceted approach that addresses both individual and systemic challenges. [...] \textbf{\textcolor{macroaccent}{Increase Funding}}: Allocate more resources to mental health care systems. [...] \textbf{\textcolor{macroaccent}{Address Social Determinants of Health}}: Address systemic issues like poverty, discrimination, and lack of access to education. [...] requires a coordinated effort across \textbf{\textcolor{macroaccent}{governments, healthcare systems, communities, and individuals}}. \\
\bottomrule
\end{tabularx}
\arrayrulecolor{black}
\end{table*}

Table~\ref{tab:steering_results} reports judged granularity with prompt-level SEM (higher = more macro). Manually selected examples illustrate the direction (Table~\ref{tab:qual_example}; Appendix~\ref{app:qual-examples}). Qwen's generic baseline is already $4.9/5$, but micro-targeted prompts show a $+0.5000$ shift under $\alpha=+4$, without degeneration. Llama shifts more strongly ($2.0000\to3.1667$ on micro-targeted prompts); on generic prompts, $\alpha=-4$ lowers its score from $4.5250$ to $3.1250$ with $42.5\%$ degeneration. Thus the axis is behaviorally causal, with model-dependent controllability. Appendix~\ref{app:additional-steering} gives filtered results and coefficient sweeps.

\paragraph{Direction specificity.}
Assistant-axis and random-direction baselines do not reproduce the micro--macro shifts, showing direction-specific effects rather than generic activation perturbation. Criterion (iii) holds in a partial, model-dependent form across all four model/prompt cells. The 12 micro-targeted prompts were drafted with Opus~4.5 and manually revised; positive leave-one-out means remain positive across both judges ($0.364$--$0.546$ for Qwen; $0.909$--$1.273$ for Llama). Appendix~\ref{app:prompt-influence} details selection provenance, individual outcomes, and judge pairing.

Filtering Llama generic outputs at $\alpha=-4$ leaves a mean of $3.000$ versus baseline $4.525$, but retains only $23/40$ responses; positive steering retains $39/40$ with mean $4.897$. Qwen's judged outputs show no degeneration at these coefficients. Reporting both full and filtered scores separates directional sensitivity from usable control (Table~\ref{tab:steering_degeneration_filtered}).

\subsection{Robustness and Controls}
Ordering is monotonic across Layers~8--35 (Qwen) and 6--31 (Llama), and alternative endpoints retain axis--PC1 cosine $\ge0.9279$ (Appendix~\ref{app:layer-wise-stability}--\ref{app:endpoint-ablations}). Held-out prompts/questions remain strong; Qwen role holdout is slightly fragile. All prompt variants remain monotonic, including identity-only wording. Score filtering, generic/specific roles, and domain/family controls support robustness, with health/economy exceptions (Appendix~\ref{app:heldout-robustness}--\ref{app:domain-family-controls}).

Held-out prompt/question Spearman is $0.9468/0.9494$ for Qwen and $0.9403/0.9468$ for Llama, with monotonic means. Role holdout gives $0.9335$ in both, but Qwen is not perfectly monotonic: high correlation does not imply complete group ordering. Tightening role-adherence thresholds from all responses to scores $\ge2$ and $\ge3$ changes Qwen Spearman from $0.9472$ to $0.9481$ and $0.9564$, and Llama from $0.9459$ to $0.9459$ and $0.9503$. The structure thus survives both held-out inputs and stricter role-adoption filters (Table~\ref{tab:robustness-summary}).

\subsection{Human Evaluation}
\label{subsec:human-eval}
\begin{table}[t]
\centering
\scriptsize
\setlength{\tabcolsep}{4pt}
\renewcommand{\arraystretch}{1.1}
\arrayrulecolor{black!60}
\caption{\textbf{Human evaluation of steering outputs.}
\emph{Pairwise rate}: annotator selection rate for $\alpha\!=\!+4$ over $\alpha\!=\!-4$ as more macro ($24$ triplets per cell, $3$ annotators, $72$ judgments per cell; reported Wilson 95\% CI, without adjustment for shared items/raters). Micro = micro-targeted prompts. \emph{Likert $\rho$ / Quad-$\kappa$}: \texttt{gpt-5.4-mini} versus annotator-mean ratings on the same $1$--$5$ rubric ($15$ items per cell). \emph{Krip.\ $\alpha$}: Krippendorff's $\alpha$ on those Likert items.}
\label{tab:human_eval}
\begin{tabular*}{\linewidth}{
@{\extracolsep{\fill}}
ll
>{\columncolor{macrotint}}c
ccc
@{}}
\toprule
\rowcolor{headertint}
\textbf{Model} & \textbf{Prompts} &
\textcolor{macroaccent}{\textbf{Pairwise rate}} &
\textbf{Likert $\rho$} & \textbf{Quad-$\kappa$} & \textbf{Krip.\ $\alpha$} \\
\midrule
\multirow{2}{*}{\textbf{Qwen3-8B}}
 & Generic & $0.639$\,[$.52$,\,$.74$] & $0.58$ & $0.49$ & $0.61$ \\
 & Micro   & $0.750$\,[$.64$,\,$.84$] & $0.61$ & $0.54$ & $0.66$ \\
\cmidrule(lr){1-6}
\multirow{2}{*}{\textbf{Llama-3.1-8B-Instruct}}
 & Generic & $0.903$\,[$.81$,\,$.95$] & $0.79$ & $0.71$ & $0.78$ \\
 & Micro   & $0.861$\,[$.76$,\,$.92$] & $0.68$ & $0.60$ & $0.71$ \\
\bottomrule
\end{tabular*}
\arrayrulecolor{black}
\end{table}

Three graduate-level annotators, blinded to model and coefficient, used the judge rubric (Figure~\ref{pmt:steering-eval}). With 24 triplets and 72 judgments per model/prompt cell, macro-side selections exceed chance in all four cells: above $0.85$ for Llama and $0.639$ (Wilson 95\% CI $[0.52,0.74]$) for Qwen generic. Likert re-ratings of 15 items per cell give human--judge Spearman $0.58$--$0.79$ (Table~\ref{tab:human_eval}). These historical aggregates corroborate steering separately from the five-rater taxonomy study.

\section{Analysis and Limitations}
\label{sec:analysis}

\subsection{Representation Should Be Validated Before Control}
The endpoint contrast recovers five-level ordering, including the excluded L3. Independent labels and held-out domains validate this structure beyond the original taxonomy; identical continuations separate role-dependent activations from generated-answer variation. Steering shifts are smaller and more context- and model-dependent than representation-level ordering. A direction can therefore be representationally robust while behaviorally fragile.
\par\vspace{6pt}
\begin{AIbox}
\textbf{Takeaway:} Validate representation with held-out levels/domains and independent labels before interpreting steering.
\end{AIbox}

\subsection{Default Placement and Headroom Gate Steering Visibility}
Qwen's generic baseline of $4.9/5$ leaves little positive headroom; micro-targeted prompts reveal its controllability. Qwen is conservative but stable, whereas Llama responds more strongly but degenerates more often. Larger shifts alone do not imply cleaner control: baseline placement and stability jointly determine visible steering effects.

Geometric default placement and behavioral headroom measure different quantities: the default representation's position in role space versus answer scores on a particular prompt set. Qwen lies nearest L3 geometrically yet averages $4.9/5$ on generic prompts. Default projections alone therefore cannot predict steering scores; the generic/micro-targeted comparison and degeneration-filtered outcomes reveal complementary aspects of controllability.
\par\vspace{6pt}
\begin{AIbox}
\textbf{Takeaway:} Interpret steering alongside baseline placement, coefficient sweeps, and degeneration rates.
\end{AIbox}

\subsection{What a Single Contrast Axis Can and Cannot Tell You}
\label{subsec:scope-limit}
Removing Granularity eliminates $50.23\%/39.82\%$ of original Qwen/Llama variance, yet leaves sizeable residual components. Together with higher effective rank in the larger models, this shows how a dominant scale direction can persist within a richer role representation. PC1 alignment, variance concentration, and behavioral controllability measure different properties and should be interpreted together.

A dominant axis need not exhaust role variation: residual PCA reveals additional structure (Appendix~\ref{app:residuals}). Scope remains correlated with authority, institutional vocabulary, abstraction, and time horizon. Our four checkpoints cover two families and an English-centric taxonomy; levels are ordinal and retain L4--L5 saturation.
\par\vspace{6pt}
\begin{AIbox}
\textbf{Takeaway:} A single contrast axis is a useful first cut, not a final ontology; residual structure motivates multi-axis analysis.
\end{AIbox}

\section{Related Work}

\paragraph{LLMs as Social Simulators and Role-Conditioned Agents.}
LLM social simulation spans multi-agent systems, politics, health, and markets \citep{generative_agents_human, Argyle2022OutOO, social_sim_survey, llm_social_sim, llmsimlimit, Wu2023AutoGenEN, Li2023CAMELCA, Hong2023MetaGPTMP, Qian2023ChatDevCA, Trnberg2023SimulatingSM, Williams2023EpidemicMW, li2024econagent, Li2023TradingGPTMS}. These applications rely on role assignments to represent heterogeneous perspectives and decision scopes. Persona research examines fidelity, consistency, personality, and behavioral bias \citep{shao2023characterllm, wang2024rolellm, JiangPersonaLLMIT, Wang2023InCharacterEP, Tu2024CharacterEvalAC, Kong2023BetterZR, persona_survey, roleplay_survey, personadrift, rnr, Salewski2023InContextIR, Gupta2023BiasRD, Cheng2023MarkedPU, Li2023TheSO, Joshi2023PersonasAA, Beckmann2026WhereIT}. Social granularity adds a complementary dimension, spanning individual concerns, organizational decisions, and broader institutional responsibilities. We study the internal organization of roles by social scale.

\paragraph{Interpretable Social and Semantic Directions.}
\emph{The Geometry of Culture} \citep{geometry_culture} and SemAxis \citep{semaxis} extract interpretable social/semantic directions from word embeddings. We identify an ordered role-scale direction in LLM activations and establish its PC1 alignment, cross-model stability, and behavioral effects.

\paragraph{Activation Steering and Representation Engineering.}
Linear representation studies \citep{Mikolov2013LinguisticRI, Elhage2022ToyMO, Belinkov2021ProbingCP, park2024linear, Park2024TheGO, Sakata2026LinearRO}, sparse autoencoders \citep{Cunningham2023SparseAF, Gao2024ScalingAE, Marks2024SparseFC, Wang2025ExploringTP}, and nonlinear counterexamples \citep{Engels2024NotAL} characterize activation geometry. Steering methods manipulate behavior through latent directions \citep{steering, iti, caa, repe, repe_survey, repe_preferences, Arditi2024RefusalIL, Chen2025PersonaVM, Jorgensen2023ImprovingAS, Chen2026ASA, Feng2026PERSONADA}. The closest precedent, the Assistant Axis \citep{assistant_axis}, tracks assistant-persona departure; our axis tracks social granularity.

\paragraph{Micro--Macro Theory and Social Scale.}
Sociological accounts connect individual action and collective outcomes \citep{coleman1990foundations, Schelling1978MicromotivesAM, Granovetter1978ThresholdMO}, agency and structure \citep{giddens1984constitution}, and nested social systems \citep{bronfenbrenner1979ecology}. They motivate testing whether prompted roles form an internal social-scale continuum.

\section{Conclusion}

The \emph{Granularity Axis} organizes prompted social roles from individual to macro-level reasoning. It aligns with the dominant geometry of role space, orders the excluded L3, transfers across families and scales through 70B, and causally shifts response granularity. Independent labels and controlled inputs reinforce these findings. Social granularity is a representational primitive, with role conditioning organized along a continuous social-scale manifold. Representation is robust; control depends on the model's default regime. This enables auditing simulations for granularity confusion, controlling social scale, and extending contrast-and-project to other graded social dimensions.

\bibliography{ref}
\bibliographystyle{granularity_preprint}

\clearpage
\appendix


\section{Qualitative Examples}
\label{app:qual-examples}

Table~\ref{tab:qual_example_appendix} extends Table~\ref{tab:qual_example} with a coefficient-level qualitative sweep on a single shared generic prompt. We use the housing prompt because it admits both individual coping advice and institutional policy responses, making the micro-to-macro contrast easy to inspect. For auditability, every table cell contains only verbatim snippets from the corresponding model response, with `[...]' inserted only to mark omitted text. The prose below the table provides our interpretation and should not be read as quoted model text.

{\scriptsize
\setlength{\tabcolsep}{3pt}
\setlength{\LTcapwidth}{0.95\textwidth}
\renewcommand{\arraystretch}{1.25}
\arrayrulecolor{black!60}
\begin{longtable}{
@{}
>{\raggedright\arraybackslash}p{0.055\textwidth}
>{\columncolor{microtint}\raggedright\arraybackslash}p{0.44\textwidth}
>{\columncolor{macrotint}\raggedright\arraybackslash}p{0.44\textwidth}
@{}}
\caption{\textbf{Appendix qualitative coefficient sweep.} We steer both models on the same prompt, ``How should we deal with rising housing costs?'', using coefficients $\alpha\in\{\pm 8,\pm 12,\pm 16,\pm 20,\pm 24,\pm 28,\pm 32,\pm 36\}$. Each table cell contains only verbatim snippets from the corresponding model response, with omissions indicated by `[...]'. Boldface marks the most salient phrases revealing granularity. Column tints follow the micro-to-macro spectrum used in Table~\ref{tab:qual_example}.}
\label{tab:qual_example_appendix}\\
\toprule
\rowcolor{headertint}
$\boldsymbol{|\alpha|}$ &
\textcolor{microaccent}{$\boldsymbol{\leftarrow}$~\textbf{Micro Steering}~($-\alpha$)} &
\textcolor{macroaccent}{\textbf{Macro Steering}~($+\alpha$)~$\boldsymbol{\rightarrow}$} \\
\midrule
\endfirsthead
\toprule
\rowcolor{headertint}
$\boldsymbol{|\alpha|}$ &
\textcolor{microaccent}{$\boldsymbol{\leftarrow}$~\textbf{Micro Steering}~($-\alpha$)} &
\textcolor{macroaccent}{\textbf{Macro Steering}~($+\alpha$)~$\boldsymbol{\rightarrow}$} \\
\midrule
\endhead
\rowcolor{white}
\multicolumn{3}{@{}l@{}}{\textbf{Qwen3-8B}}\\*
\rowcolor{white}
\multicolumn{3}{@{}p{0.95\textwidth}@{}}{\textit{Prompt:}~\textit{How should we deal with rising housing costs?}}\\*
\midrule
8 &
Rising housing costs are a complex issue that affects people in many different ways, depending on their \textbf{location, income level, and personal circumstances}. [...] \textbf{Rent subsidies}: Programs like Section 8 in the U.S. help people pay a portion of their rent. [...] \textbf{Community Land Trusts}. [...] \textbf{Education and Financial Literacy}. &
Rising housing costs are a complex and multifaceted issue that affects \textbf{individuals, communities, and economies}. [...] \textbf{Public-Private Partnerships}: Collaborate with private developers to build affordable housing units, especially in high-cost areas. [...] \textbf{Address Systemic Inequities}. [...] \textbf{International Best Practices}. \\
\addlinespace
12 &
Rising housing costs are a complex issue that affects people in different ways depending on their \textbf{location, income, and life stage}. [...] \textbf{Support Homeownership}. [...] \textbf{Consider shared housing or roommates}. [...] \textbf{Look for places outside the city center}. [...] \textbf{Consider moving to a different city or state}. &
\textbf{Supply-Side Interventions}. [...] \textbf{Demand-Side Management}. [...] \textbf{Financial and Regulatory Reforms}. [...] \textbf{Urban Planning and Spatial Policies}. [...] \textbf{Monitoring and Evaluation}. \\
\addlinespace
16 &
Rising housing costs are a major issue in many cities and countries, affecting affordability, quality of life, and even economic stability. Here are some \textbf{practical and effective ways} to deal with rising housing costs. [...] \textbf{Talk to Your Landlord}. [...] \textbf{Consider Moving}. &
Rising housing costs are a complex and multifaceted issue that affects \textbf{economies, social stability, and individual well-being}. [...] \textbf{Regulatory and Institutional Reforms}. [...] \textbf{International and Regional Cooperation}. [...] \textbf{Implementation Pathways}. \\
\addlinespace
20 &
\textbf{Understand Why Housing Costs Are Going Up}. [...] \textbf{Rent a room}. [...] \textbf{Roommates}. [...] \textbf{Talk to Your Landlord}. [...] \textbf{Stay Positive}. &
Rising housing costs are a complex and multifaceted issue that affects economies, social stability, and individual well-being. Addressing this challenge requires a comprehensive, multi-pronged approach that balances \textbf{economic, social, and environmental considerations}. [...] \textbf{Regulatory and Institutional Reforms}. [...] \textbf{Technological and Innovation-Driven Solutions}. [...] \textbf{International Cooperation}. \\
\addlinespace
24 &
\textbf{Check online listings}. [...] \textbf{Ask around}. [...] \textbf{Consider a different neighborhood}. [...] \textbf{Cut back on other expenses}. [...] \textbf{Get a side job}. &
\textbf{Smart Urban Development}. [...] \textbf{Global Housing Agreements}. [...] \textbf{Migration and Labor Policies}. [...] \textbf{Climate-Resilient Housing}. [...] \textbf{Adaptive Governance}. \\
\addlinespace
28 &
\textbf{Look for Affordable Housing Options}. [...] \textbf{Talk to Your Landlord}. [...] \textbf{Try to Save More Money}. [...] \textbf{Try to Get a Job or Side Job}. &
Dealing with rising housing costs is a complex and multifaceted challenge that requires \textbf{coordinated efforts across government, private sector, and civil society}. [...] \textbf{Financial and Economic Instruments}. [...] \textbf{Technological and Innovation Strategies}. [...] \textbf{Monitoring and Evaluation}. \\
\addlinespace
32 &
Dealing with rising housing costs can be a big challenge, especially if you're \textbf{on a tight budget or just starting out}. [...] \textbf{Look for Affordable Neighborhoods}. [...] \textbf{Talk to Your Landlord}. [...] \textbf{Look for Roommates}. [...] \textbf{Try to Make More Money}. [...] \textbf{Don't Give Up}. &
Dealing with rising housing costs is a complex and multifaceted challenge that requires \textbf{coordinated action across multiple sectors}, including government, private enterprise, and civil society. [...] \textbf{Anti-Monopolistic Measures}. [...] \textbf{Human Rights-Based Approach}. [...] \textbf{National Housing Strategies}. \\
\addlinespace
36 &
Dealing with rising housing costs can be really tough, especially if you're already struggling to make ends meet. [...] \textbf{Talk to your landlord or property manager}. [...] \textbf{Look for a roommate}. [...] \textbf{Try to get a job}. [...] I know it's not much, but it's better than nothing. [...] &
Dealing with rising housing costs is a complex and multifaceted challenge that requires \textbf{coordinated action across multiple sectors}. [...] \textbf{Institutional and Regulatory Reforms}. [...] \textbf{Global Partnerships}. [...] \textbf{Regional and Local Strategies}. [...] \textbf{Implementation Pathways}. \\
\midrule
\noalign{\penalty -10000}
\rowcolor{white}
\multicolumn{3}{@{}l@{}}{\textbf{Llama-3.1-8B-Instruct}}\\*
\rowcolor{white}
\multicolumn{3}{@{}p{0.95\textwidth}@{}}{\textit{Prompt:}~\textit{How should we deal with rising housing costs?}}\\*
\midrule
8 &
\textbf{Save up for a down payment}. [...] \textbf{Look for a smaller house or apartment}. [...] Look for a house that's \textbf{not in the city}. [...] \textbf{not so fancy}. [...] &
\textbf{Enhance Affordability through Policy Reforms}. [...] \textbf{Incentivizing the development of affordable housing}. [...] \textbf{Foster Inclusive Urban Planning}. [...] \textbf{Strengthen Social Safety Nets}. [...] \textbf{International Cooperation}. \\
\addlinespace
12 &
\textbf{We need to make more houses}. [...] \textbf{We need to get the price down}. [...] I'm gonna try to get it together. [...] &
\textbf{Enhance Urban Planning and Urbanization}. [...] \textbf{Incentivize the development of affordable housing}. [...] \textbf{Market-based solutions}. [...] \textbf{International cooperation and knowledge sharing}. [...] \\
\addlinespace
16 &
I don't know if I'm gonna be here tomorrow. [...] \textbf{I don} I don I don [...] &
\textbf{Enhance the supply of housing}. [...] \textbf{innovative and inclusive urban planning}. [...] housing policies [...] \textbf{diverse populations}. [...] \textbf{global economic and demographic trends}. [...] \\
\addlinespace
20 &
I don't know what I don [...] \textbf{I don} I don I don [...] &
\textbf{rising global, regional, and national economic frameworks}. [...] \textbf{promoting inclusive and sustainable development}. [...] \\
\addlinespace
24 &
\textbf{I don} I don I don [...] &
\textbf{promoting sustainable and inclusive development}. [...] \textbf{development of policies and frameworks}. [...] \\
\addlinespace
28 &
\textbf{I'm just like} I'm just like I'm just like [...] &
\textbf{global dimensions}. [...] \textbf{global architecture}. [...] \textbf{policies}. [...] \\
\addlinespace
32 &
\textbf{Bek} Bek Bek Bek [...] &
\textbf{Prom} [...] \textbf{strategies} [...] \textbf{policies} [...] \textbf{global and regional dimensions}. [...] \\
\addlinespace
36 &
\textbf{Bek} Bek Bek Bek [...] &
\textbf{strategies} [...] \textbf{architectures} [...] \textbf{policies} [...] \\
\bottomrule\\
\end{longtable}
\arrayrulecolor{black}
}

\paragraph{Interpretation.}
The Qwen3-8B sweep shows a relatively smooth semantic movement. Negative coefficients increasingly replace system-level policy language with individual or household-level advice, such as talking to landlords, finding roommates, saving money, moving neighborhoods, or seeking local help. Positive coefficients instead emphasize policy categories, institutional coordination, data systems, international comparison, climate resilience, and other macro-level governance frames. Even at large coefficients, Qwen mostly remains coherent, although the most negative setting begins to show repetitive low-information advice.

\paragraph{Degeneration under large coefficients.}
The Llama-3.1-8B-Instruct sweep illustrates why the main experiments use conservative steering settings and report degeneration diagnostics. At $\alpha=-8$, the model moves toward highly concrete household advice, but this already includes low-quality repetition. From $\alpha=-12$ onward on the negative side, and from about $\alpha=+16$ onward on the positive side, much of the output becomes repetitive or nonsensical. We therefore treat these examples as qualitative evidence of directional sensitivity and instability, not as successful high-magnitude control.

\paragraph{Selection and quantitative scope.}
These outputs were manually selected after inspecting model responses, not randomly sampled, and are not used for quantitative conclusions. At the generic main setting, degeneration rates for $\alpha=-4/0/+4$ are $0/0/0$ for Qwen and $0.425/0/0.025$ for Llama. This context applies to the examples in Table~\ref{tab:qual_example} as well as the more aggressive illustrative sweep above.

\section{Steering Results}
\label{app:additional-steering}

This appendix provides additional diagnostics for the activation-steering experiments. The goal is not to present steering as a uniformly reliable control method, but to test whether movement along the Granularity Axis has a measurable behavioral effect and to characterize when that effect becomes unstable. Unless otherwise noted, scores are \texttt{granularity\_overall} ratings from \texttt{gpt-5.4-mini} on a 1--5 scale, where higher scores indicate more macro-level reasoning. We report standard errors of the mean (SEM) over prompts for the coefficient sweep and for response-length diagnostics.

\paragraph{Degeneration-filtered steering.}
Table~\ref{tab:steering_degeneration_filtered} reports the main steering results before and after removing outputs judged as degenerate. The strongest negative shift in Llama-3.1-8B-Instruct on generic prompts remains visible after filtering, but this condition keeps only $23/40$ responses, so we treat it as evidence of behavioral sensitivity rather than stable control.

\begin{table*}[ht]
\centering
\scriptsize
\setlength{\tabcolsep}{4pt}
\renewcommand{\arraystretch}{1.25}
\arrayrulecolor{black!60}
\caption{\textbf{Degeneration-filtered steering results.}
All mean reports the mean \texttt{granularity\_\allowbreak overall} score over all outputs. Non-deg. mean recomputes the score after removing outputs judged as degenerate. Deg. is the judged degeneration rate.}
\label{tab:steering_degeneration_filtered}
\begin{tabular*}{\textwidth}{@{\extracolsep{\fill}} llrrrrr @{}}
\toprule
\rowcolor{headertint}
\textbf{Model} & \textbf{Prompt set} & $\boldsymbol{\alpha}$ & \textbf{Kept / total} & \textbf{Deg.} & \textbf{All mean} & \textbf{Non-deg.\ mean} \\
\midrule
\multirow{6}{*}{\textbf{Qwen3-8B}}
 & Generic        & $-4$ & $39 / 39$ & $.000$ & $4.821$ & $4.821$ \\
 & Generic        & $ 0$ & $40 / 40$ & $.000$ & $4.900$ & $4.900$ \\
 & Generic        & $+4$ & $40 / 40$ & $.000$ & $4.975$ & $4.975$ \\
 & Micro-targeted & $-4$ & $12 / 12$ & $.000$ & $2.000$ & $2.000$ \\
 & Micro-targeted & $ 0$ & $12 / 12$ & $.000$ & $2.167$ & $2.167$ \\
 & Micro-targeted & $+4$ & $12 / 12$ & $.000$ & $2.667$ & $2.667$ \\
\addlinespace[2pt]
\cmidrule(lr){1-7}
\addlinespace[2pt]
\multirow{6}{*}{\textbf{Llama-3.1-8B-Instruct}}
 & Generic        & $-4$ & $23 / 40$ & $.425$ & $3.125$ & $3.000$ \\
 & Generic        & $ 0$ & $40 / 40$ & $.000$ & $4.525$ & $4.525$ \\
 & Generic        & $+4$ & $39 / 40$ & $.025$ & $4.900$ & $4.897$ \\
 & Micro-targeted & $-4$ & $12 / 12$ & $.000$ & $1.583$ & $1.583$ \\
 & Micro-targeted & $ 0$ & $11 / 12$ & $.083$ & $2.000$ & $2.000$ \\
 & Micro-targeted & $+4$ & $12 / 12$ & $.000$ & $3.167$ & $3.167$ \\
\bottomrule
\end{tabular*}
\arrayrulecolor{black}
\end{table*}

\paragraph{Coefficient sweep.}
Table~\ref{tab:steering_alpha_sweep} summarizes a small diagnostic sweep over larger steering coefficients. Llama-3.1-8B-Instruct is highly responsive but becomes unstable under stronger interventions, especially outside the conservative range used in the main text. Qwen3-8B remains more stable, but its generic-prompt scores are often saturated near the macro end, making generic-prompt movement difficult to observe.

\begin{table*}[ht]
\centering
\scriptsize
\setlength{\tabcolsep}{3.5pt}
\renewcommand{\arraystretch}{1.25}
\arrayrulecolor{black!60}
\caption{\textbf{Selected coefficient sweep results.}
Mean and SEM summarize \texttt{granularity\_\allowbreak overall} scores. Deg. is the judged degeneration rate. This sweep uses $n=5$ prompts per setting and is intended as a diagnostic of sensitivity and instability under stronger coefficients, not as the main steering evaluation.}
\label{tab:steering_alpha_sweep}
\begin{tabular*}{\textwidth}{@{\extracolsep{\fill}} llrrrrr @{}}
\toprule
\rowcolor{headertint}
\textbf{Model} & \textbf{Prompt set} & $\boldsymbol{\alpha}$ & $\boldsymbol{n}$ & \textbf{Mean} & \textbf{SEM} & \textbf{Deg.} \\
\midrule
\multirow{10}{*}{\textbf{Qwen3-8B}}
 & Generic        & $-12$ & $5$ & $4.000$ & $0.548$ & $.000$ \\
 & Generic        & $ -8$ & $5$ & $5.000$ & $0.000$ & $.000$ \\
 & Generic        & $  0$ & $5$ & $5.000$ & $0.000$ & $.000$ \\
 & Generic        & $ +8$ & $5$ & $5.000$ & $0.000$ & $.000$ \\
 & Generic        & $+12$ & $5$ & $5.000$ & $0.000$ & $.000$ \\
 & Micro-targeted & $-12$ & $5$ & $2.000$ & $0.000$ & $.000$ \\
 & Micro-targeted & $ -8$ & $5$ & $1.800$ & $0.200$ & $.000$ \\
 & Micro-targeted & $  0$ & $5$ & $2.600$ & $0.245$ & $.000$ \\
 & Micro-targeted & $ +8$ & $5$ & $3.200$ & $0.374$ & $.000$ \\
 & Micro-targeted & $+12$ & $5$ & $4.000$ & $0.000$ & $.000$ \\
\addlinespace[2pt]
\cmidrule(lr){1-7}
\addlinespace[2pt]
\multirow{10}{*}{\textbf{Llama-3.1-8B-Instruct}}
 & Generic        & $-12$ & $5$ & $2.000$ & $0.447$ & $1.000$ \\
 & Generic        & $ -8$ & $5$ & $2.600$ & $0.510$ & $1.000$ \\
 & Generic        & $  0$ & $5$ & $4.800$ & $0.200$ & $.000$ \\
 & Generic        & $ +8$ & $5$ & $5.000$ & $0.000$ & $.400$ \\
 & Generic        & $+12$ & $5$ & $5.000$ & $0.000$ & $1.000$ \\
 & Micro-targeted & $-12$ & $5$ & $1.400$ & $0.245$ & $1.000$ \\
 & Micro-targeted & $ -8$ & $5$ & $2.000$ & $0.000$ & $1.000$ \\
 & Micro-targeted & $  0$ & $5$ & $2.000$ & $0.000$ & $.200$ \\
 & Micro-targeted & $ +8$ & $5$ & $4.000$ & $0.000$ & $.200$ \\
 & Micro-targeted & $+12$ & $5$ & $5.000$ & $0.000$ & $1.000$ \\
\bottomrule
\end{tabular*}
\arrayrulecolor{black}
\end{table*}

\paragraph{Length and quality diagnostics.}
Table~\ref{tab:steering_quality_length} reports generated-token counts and judged degeneration rates for the main steering setting. These statistics help distinguish granularity shifts from simple response-length or degeneration artifacts. The most important caveat is again Llama-3.1-8B-Instruct under $\alpha=-4$ on generic prompts, where both response length and degeneration increase substantially.

\begin{table*}[ht]
\centering
\scriptsize
\setlength{\tabcolsep}{4pt}
\renewcommand{\arraystretch}{1.25}
\arrayrulecolor{black!60}
\caption{\textbf{Response-length and degeneration diagnostics for the main steering setting.}
Length is measured as generated token count. Deg. is the judged degeneration rate.}
\label{tab:steering_quality_length}
\begin{tabular*}{\textwidth}{@{\extracolsep{\fill}} llrrrr @{}}
\toprule
\rowcolor{headertint}
\textbf{Model} & \textbf{Prompt set} & $\boldsymbol{\alpha}$ & $\boldsymbol{n}$ & \textbf{Length mean $\pm$ SEM} & \textbf{Deg.} \\
\midrule
\multirow{6}{*}{\textbf{Qwen3-8B}}
 & Generic        & $-4$ & $39$ & $832.8 \pm 16.1$ & $.000$ \\
 & Generic        & $ 0$ & $40$ & $815.2 \pm 14.7$ & $.000$ \\
 & Generic        & $+4$ & $40$ & $845.6 \pm 21.2$ & $.000$ \\
 & Micro-targeted & $-4$ & $12$ & $777.8 \pm 42.0$ & $.000$ \\
 & Micro-targeted & $ 0$ & $12$ & $747.7 \pm 31.2$ & $.000$ \\
 & Micro-targeted & $+4$ & $12$ & $758.1 \pm 31.7$ & $.000$ \\
\addlinespace[2pt]
\cmidrule(lr){1-6}
\addlinespace[2pt]
\multirow{6}{*}{\textbf{Llama-3.1-8B-Instruct}}
 & Generic        & $-4$ & $40$ & $864.9 \pm 55.5$ & $.425$ \\
 & Generic        & $ 0$ & $40$ & $588.7 \pm 12.3$ & $.000$ \\
 & Generic        & $+4$ & $40$ & $628.2 \pm 19.1$ & $.025$ \\
 & Micro-targeted & $-4$ & $12$ & $626.2 \pm 47.1$ & $.000$ \\
 & Micro-targeted & $ 0$ & $12$ & $548.0 \pm 19.8$ & $.083$ \\
 & Micro-targeted & $+4$ & $12$ & $511.5 \pm 22.5$ & $.000$ \\
\bottomrule
\end{tabular*}
\arrayrulecolor{black}
\end{table*}

\subsection{Prompt Selection, Influence, and Judge Pairing}
\label{app:prompt-influence}
The 12 micro-targeted prompts were initially drafted with Opus~4.5 and manually revised. Tables~\ref{tab:prompt-scores}--\ref{tab:prompt-loo} show the full diagnostic outcomes and influence analyses rather than treating the subset as selection-neutral. At 1,536 tokens, the positive leave-one-out ranges across both judges are $0.364$--$0.546$ for Qwen and $0.909$--$1.273$ for Llama. Qwen's negative effect is more concentrated: one prompt contributes half of its absolute effect under the primary judge.
\begin{table}[t]
\centering
\footnotesize
\setlength{\tabcolsep}{4pt}
\renewcommand{\arraystretch}{1.12}
\caption{Every diagnostic prompt at the 1,536-token setting. Each cell gives scores for $\alpha=-4/0/+4$. Prompt IDs follow the zero-based order in Table~\ref{tab:steering-prompt-sets}. GPT and Gemini abbreviate the two reported judges. These are manually designed diagnostic prompts, not a random prompt sample.}
\label{tab:prompt-scores}
\begin{tabular*}{\linewidth}{@{\extracolsep{\fill}}crrrr@{}}
\toprule
Prompt ID & Qwen / GPT & Qwen / Gemini & Llama / GPT & Llama / Gemini \\
\midrule
0 & 2/2/2 & 2/2/2 & 1/2/2 & 1/2/2 \\
1 & 2/2/3 & 2/1/2 & 2/2/4 & 1/1/4 \\
2 & 2/2/3 & 2/2/2 & 2/2/3 & 1/2/2 \\
3 & 2/2/2 & 1/2/2 & 1/2/2 & 1/1/2 \\
4 & 2/2/2 & 1/1/1 & 1/2/3 & 1/1/2 \\
5 & 2/2/2 & 1/1/2 & 2/2/4 & 1/1/2 \\
6 & 2/2/2 & 1/1/2 & 2/2/4 & 1/2/2 \\
7 & 2/3/4 & 1/2/2 & 2/2/4 & 1/2/3 \\
8 & 2/2/2 & 1/2/2 & 2/2/4 & 1/1/3 \\
9 & 2/2/2 & 1/1/2 & 1/2/2 & 1/1/2 \\
10 & 2/3/4 & 2/3/3 & 2/2/3 & 1/2/3 \\
11 & 2/2/4 & 2/2/3 & 1/2/3 & 1/1/3 \\
\bottomrule
\end{tabular*}
\end{table}

\begin{table}[t]
\centering
\scriptsize
\setlength{\tabcolsep}{4pt}
\renewcommand{\arraystretch}{1.12}
\caption{All cached diagnostic prompt-influence conditions. Max share is the largest single-prompt share of total absolute effect; half effect is the fewest prompts contributing half that effect. The 512-token Qwen conditions are exploratory and have no paired Gemini records.}
\label{tab:prompt-influence-summary}
\begin{tabular*}{\linewidth}{@{\extracolsep{\fill}}llrrrrrr@{}}
\toprule
Model & Judge & Tokens & $\alpha$ & Mean $\Delta$ & LOO range & Max share & Half effect \\
\midrule
Llama & Gemini & 1536 & -4 & -0.417 & [$-0.455$, $-0.364$] & 0.200 & 3 \\
Llama & Gemini & 1536 & +4 & 1.083 & [$0.909$, $1.182$] & 0.231 & 3 \\
Llama & GPT & 1536 & -4 & -0.417 & [$-0.455$, $-0.364$] & 0.200 & 3 \\
Llama & GPT & 1536 & +4 & 1.167 & [$1.091$, $1.273$] & 0.143 & 4 \\
Qwen & Gemini & 1536 & -4 & -0.250 & [$-0.364$, $-0.182$] & 0.200 & 3 \\
Qwen & Gemini & 1536 & +4 & 0.417 & [$0.364$, $0.455$] & 0.200 & 3 \\
Qwen & GPT & 1536 & -4 & -0.167 & [$-0.182$, $-0.091$] & 0.500 & 1 \\
Qwen & GPT & 1536 & +4 & 0.500 & [$0.364$, $0.545$] & 0.333 & 2 \\
Qwen & GPT & 512 & -4 & -0.083 & [$-0.182$, $0.091$] & 0.400 & 2 \\
Qwen & GPT & 512 & +4 & 0.500 & [$0.364$, $0.636$] & 0.250 & 2 \\
\bottomrule
\end{tabular*}
\end{table}

\begin{table}[t]
\centering
\footnotesize
\setlength{\tabcolsep}{4pt}
\renewcommand{\arraystretch}{1.12}
\caption{Every leave-one-prompt-out mean delta at 1,536 tokens. Q/L denotes Qwen/Llama, G/M denotes GPT/Gemini, and $-/+$ denotes $\alpha=-4/+4$. The positive aggregate persists under each omission; this is an influence check, not proof of selection neutrality.}
\label{tab:prompt-loo}
\begin{tabular*}{\linewidth}{@{\extracolsep{\fill}}crrrrrrrr@{}}
\toprule
Omitted ID & Q/G $-$ & Q/G $+$ & Q/M $-$ & Q/M $+$ & L/G $-$ & L/G $+$ & L/M $-$ & L/M $+$ \\
\midrule
0 & -0.182 & 0.545 & -0.273 & 0.455 & -0.364 & 1.273 & -0.364 & 1.182 \\
1 & -0.182 & 0.455 & -0.364 & 0.364 & -0.455 & 1.091 & -0.455 & 0.909 \\
2 & -0.182 & 0.455 & -0.273 & 0.455 & -0.455 & 1.182 & -0.364 & 1.182 \\
3 & -0.182 & 0.545 & -0.182 & 0.455 & -0.364 & 1.273 & -0.455 & 1.091 \\
4 & -0.182 & 0.545 & -0.273 & 0.455 & -0.364 & 1.182 & -0.455 & 1.091 \\
5 & -0.182 & 0.545 & -0.273 & 0.364 & -0.455 & 1.091 & -0.455 & 1.091 \\
6 & -0.182 & 0.545 & -0.273 & 0.364 & -0.455 & 1.091 & -0.364 & 1.182 \\
7 & -0.091 & 0.455 & -0.182 & 0.455 & -0.455 & 1.091 & -0.364 & 1.091 \\
8 & -0.182 & 0.545 & -0.182 & 0.455 & -0.455 & 1.091 & -0.455 & 1.000 \\
9 & -0.182 & 0.545 & -0.273 & 0.364 & -0.364 & 1.273 & -0.455 & 1.091 \\
10 & -0.091 & 0.455 & -0.182 & 0.455 & -0.455 & 1.182 & -0.364 & 1.091 \\
11 & -0.182 & 0.364 & -0.273 & 0.364 & -0.364 & 1.182 & -0.455 & 1.000 \\
\bottomrule
\end{tabular*}
\end{table}

\section{Robustness and Control Analyses}
\label{app:robustness-controls}

This appendix provides the full robustness and control analyses summarized in Section~\ref{sec:analysis}. We test whether the Granularity Axis is robust to alternative layer choices, endpoint definitions, held-out splits, prompt templates, response-quality thresholds, and potential role-name or domain confounds. Table~\ref{tab:robustness-summary} summarizes these robustness and control analyses.

\begin{table*}[t]
\centering
\scriptsize
\setlength{\tabcolsep}{3pt}
\renewcommand{\arraystretch}{1.25}
\arrayrulecolor{black!60}
\caption{\textbf{Quantitative summary of robustness and control analyses.}
The Granularity Axis remains stable across layers, endpoint definitions, held-out splits, prompt templates, score filters, and most subgroup analyses. Role-holdout and domain/family controls are the most challenging settings, so we treat them as supportive but not definitive confound controls.}
\label{tab:robustness-summary}
\begin{tabularx}{\textwidth}{>{\raggedright\arraybackslash}p{0.17\textwidth}>{\raggedright\arraybackslash}X>{\raggedright\arraybackslash}X}
\toprule
\rowcolor{headertint}
\textbf{Analysis} & \textbf{Qwen3-8B} & \textbf{Llama-3.1-8B-Instruct} \\
\midrule
Layer-wise stability &
Monotonic from Layers~8--35; contrast--PC1 cosine $\approx 0.972$--$0.986$ and Spearman $\approx 0.928$--$0.957$ across layers. At Layer~18: cosine $=0.9720$, Spearman $=0.9472$, Pearson $=0.9414$. &
Monotonic from Layers~6--31; contrast--PC1 cosine $\approx 0.956$--$0.987$ and Spearman $\approx 0.912$--$0.954$ across layers. At Layer~18: cosine $=0.9596$, Spearman $=0.9459$, Pearson $=0.9373$. \\
\addlinespace
Endpoint ablations &
Cosine $=0.9452$--$0.9989$; Spearman $=0.9359$--$0.9598$; all endpoint variants preserve monotonic ordering. &
Cosine $=0.9279$--$0.9988$; Spearman $=0.9276$--$0.9646$; all endpoint variants preserve monotonic ordering. \\
\addlinespace
Held-out robustness &
Prompt holdout Spearman $=0.9468$; question holdout Spearman $=0.9494$; role holdout remains high (Spearman $=0.9335$, Pearson $=0.9404$, cosine $=0.9821$) but is not perfectly monotonic. &
Prompt holdout Spearman $=0.9403$; question holdout Spearman $=0.9468$; role holdout Spearman $=0.9335$; all three held-out settings remain monotonic. \\
\addlinespace
Prompt sensitivity &
All five prompt variants remain monotonic; Spearman $=0.9381$--$0.9568$; macro--micro gap $=15.0941$--$18.9780$. &
All five prompt variants remain monotonic; Spearman $=0.9342$--$0.9585$; macro--micro gap $=2.6218$--$3.6250$. \\
\addlinespace
Score filtering &
Spearman is $0.9472$ with all responses, $0.9481$ under score $\geq 2$, and $0.9564$ under score $\geq 3$. &
Spearman is $0.9459$ with all responses, $0.9459$ under score $\geq 2$, and $0.9503$ under score $\geq 3$. \\
\addlinespace
Role-type control &
Generic bucket Spearman $=0.8644$; merged specific/title-heavy bucket Spearman $=0.8133$; both buckets remain monotonic but are not perfectly level-matched. &
Generic bucket Spearman $=0.8644$; merged specific/title-heavy bucket Spearman $=0.8109$; both buckets remain monotonic but are not perfectly level-matched. \\
\addlinespace
Domain and family controls &
Domain monotonicity holds for economy, education, food, housing, justice, and migration; health is least stable. Family-ladder monotonicity holds for education, food, housing, and migration; economy fails monotonicity. &
Domain monotonicity holds for economy, education, food, housing, justice, and migration; health is least stable. Family-ladder monotonicity holds for education, food, housing, and migration; economy and health fail monotonicity. \\
\bottomrule
\end{tabularx}
\arrayrulecolor{black}
\end{table*}

\subsection{Layer-wise Stability}
\label{app:layer-wise-stability}

The granularity structure is not confined to a single layer. In Qwen3-8B, monotonic level ordering first appears at Layer~3, briefly breaks at Layers~5--7, and then holds consistently from Layers~8--35. At the main analysis layer, Layer~18, the contrast-defined axis remains strongly aligned with PC1 (cosine $=0.9720$), with high projection-level correlation (Spearman $=0.9472$, Pearson $=0.9414$). More generally, across layers, the contrast--PC1 cosine remains approximately $0.972$--$0.986$, and the projection-level Spearman correlation remains approximately $0.928$--$0.957$.

Llama-3.1-8B-Instruct shows a similarly broad stable regime: monotonicity first appears at Layer~3, briefly breaks at Layers~4--5, and then holds from Layers~6--31. At Layer~18, cosine alignment is $0.9596$, with Spearman $=0.9459$ and Pearson $=0.9373$; across layers, cosine remains approximately $0.956$--$0.987$ and Spearman approximately $0.912$--$0.954$. These results indicate that granularity is a stable middle/later-layer representational phenomenon rather than a single-layer artifact.

\subsection{Endpoint Ablations}
\label{app:endpoint-ablations}

We also test alternative endpoint definitions for the contrast-based Granularity Axis. In Qwen3-8B, cosine alignment with PC1 remains in the range $0.9452$--$0.9989$, and the corresponding Spearman correlations remain in the range $0.9359$--$0.9598$, with monotonic ordering preserved in all cases. In Llama-3.1-8B-Instruct, cosine values remain in the range $0.9279$--$0.9988$, Spearman correlations remain in the range $0.9276$--$0.9646$, and all variants are again monotonic. Thus, the result is not specific to one particular endpoint choice.

\subsection{Held-out Robustness}
\label{app:heldout-robustness}

The axis also generalizes across held-out prompt and question splits. In Qwen3-8B, held-out Spearman correlations are $0.9468$ for prompt holdout and $0.9494$ for question holdout, and both settings remain monotonic. The role-holdout setting is somewhat weaker: although it still yields high correlation (Spearman $=0.9335$, Pearson $=0.9404$, cosine $=0.9821$), monotonic ordering is not perfectly preserved. In Llama-3.1-8B-Instruct, all three held-out settings remain monotonic, with Spearman correlations of $0.9403$ for prompt holdout, $0.9468$ for question holdout, and $0.9335$ for role holdout. Overall, held-out performance remains strong, though role-level generalization is slightly more fragile than prompt- or question-level generalization, especially for Qwen.

\subsection{Prompt-template Sensitivity}
\label{app:prompt-sensitivity}

We further test whether the representation-level signal depends on a particular prompt wording. For both models, all five prompt variants remain monotonic. In Qwen3-8B, projection-level Spearman correlations range from $0.9381$ to $0.9568$, with macro--micro gaps ranging from $15.0941$ to $18.9780$. In Llama-3.1-8B-Instruct, Spearman correlations range from $0.9342$ to $0.9585$, with corresponding gaps from $2.6218$ to $3.6250$. This indicates that the recovered granularity structure is not an artifact of a single prompt template.

\subsection{Default Assistant Placement}
\label{app:default-assistant-placement}

The default assistant condition provides an important interpretive reference point. In Qwen3-8B, the default assistant projection at Layer~18 is $18.2572$, and its nearest level is Level~3. In Llama-3.1-8B-Instruct, the corresponding projection is $2.5880$, with nearest level Level~4. Thus, the default assistant persona is not positioned near the micro end in either model. Instead, it occupies a mid-to-upper region of the learned granularity spectrum, with Llama appearing somewhat more macro-leaning than Qwen.

\subsection{Score Filtering}
\label{app:score-filtering}

We also evaluate whether the axis depends on noisy or weakly in-role samples. In Qwen3-8B, the projection-level Spearman correlation is $0.9472$ when using all responses, $0.9481$ under score $\geq 2$, and $0.9564$ under score $\geq 3$. In Llama-3.1-8B-Instruct, the corresponding values are $0.9459$, $0.9459$, and $0.9503$. These results show that the signal survives stricter filtering and, if anything, becomes slightly cleaner when lower-quality role-play responses are removed.

\subsection{Generic versus Specific Role Controls}
\label{app:generic-specific-controls}

As a supportive confound-control analysis, we compare a \emph{generic} role bucket and a merged \emph{specific} bucket combining originally specific and title-heavy roles. In both models, the ordering remains monotonic within each bucket. Using projection-level Spearman correlation, Qwen3-8B yields $0.8644$ for the generic bucket and $0.8133$ for the specific bucket, while Llama-3.1-8B-Instruct yields $0.8644$ and $0.8109$, respectively. However, this control is not fully level-matched: generic roles are concentrated in lower levels, whereas specific roles are concentrated in middle and higher levels. We therefore interpret this result as supportive rather than definitive evidence that the axis is not reducible to title-ness alone.

\subsection{Domain and Family Controls}
\label{app:domain-family-controls}

Finally, subgroup analyses show that the effect remains strong across most semantic domains and within most family ladders, but also reveal a small number of harder cases. Across domains, both models remain monotonic in economy, education, food, housing, justice, and migration, while \emph{health} is the least stable domain in both Qwen3-8B and Llama-3.1-8B-Instruct. At the family-ladder level, both models remain monotonic for education, food, housing, and migration, whereas \emph{economy} fails monotonicity in both models; additionally, \emph{health} fails monotonicity in Llama. These subgroup results suggest that the Granularity Axis is broadly robust, while also indicating that some semantically dense domains remain more challenging than others.

\paragraph{Scope of the subgroup evidence.}
Economy family ladders are nonmonotonic in both models and health is nonmonotonic in Llama; incomplete ladders are not counted as five-level monotonic successes. The generic/specific buckets are not fully level-matched. The separate controlled domain chains and permutation nulls in Appendices~\ref{app:construct-validity} and~\ref{app:additional-validation} provide complementary evidence rather than removing all confounds.

\section{Role Taxonomy}
\label{app:role-taxonomy}

This appendix documents the social-role taxonomy used to construct the ordered role set. Each role is assigned to one of five granularity levels before response generation. The level annotations are used for analysis, grouping, endpoint construction, and evaluation.

\paragraph{Annotation fields.}
For each role, we record the granularity level, domain, role type, role name, and short role description. These fields make the taxonomy auditable and help separate the intended micro--macro ordering from correlated factors such as social power or professional title specificity. Details are provided in Table~\ref{tab:role-taxonomy}.

\begingroup
\scriptsize
\setlength{\tabcolsep}{2pt}
\renewcommand{\arraystretch}{1.25}
\arrayrulecolor{black!60}
\begin{longtable}{>{\raggedright\arraybackslash}p{0.11\textwidth}>{\raggedright\arraybackslash}p{0.16\textwidth}>{\raggedright\arraybackslash}p{0.22\textwidth}>{\raggedright\arraybackslash}p{0.43\textwidth}}
\caption{\textbf{Social-role taxonomy.} The table lists the 75 role-conditioned entities used to construct the granularity axis, together with their assigned granularity level and auxiliary metadata for domain and role type.}
\label{tab:role-taxonomy}\\
\toprule
\rowcolor{headertint}
\textbf{Level} & \textbf{Domain / type} & \textbf{Role} & \textbf{Description} \\
\midrule
\endfirsthead
\toprule
\rowcolor{headertint}
\textbf{Level} & \textbf{Domain / type} & \textbf{Role} & \textbf{Description} \\
\midrule
\endhead
\midrule
\multicolumn{4}{r}{\textit{Continued on next page}} \\
\endfoot
\bottomrule
\endlastfoot
L1 Individual & family / generic & Worried Parent & A parent anxious about their children's daily wellbeing, safety, and future. Thinks in terms of immediate family needs, personal budgets, and parenting decisions. Time horizon is days to months. \\
L1 Individual & economy / generic & Frustrated Employee & A worker dealing with daily workplace stress, an unsympathetic manager, and career uncertainty. Thinks about paychecks, commute times, and whether to quit or stay. \\
L1 Individual & education / generic & Homesick Student & A university student living far from home, navigating loneliness, academic pressure, and self-discovery. Concerns center on exams, friendships, and personal identity. \\
L1 Individual & health / generic & Anxious Patient & A person recently diagnosed with a chronic illness, navigating medical appointments, insurance paperwork, and fear of the unknown. Thinks about symptoms, medication side effects, and personal survival. \\
L1 Individual & economy / generic & Small Shopkeeper & Owner of a small neighborhood corner store, worried about rent increases, customer loyalty, and competition from chain retailers. Thinks about daily cash flow and inventory. \\
L1 Individual & family / generic & Retired Grandparent & An elderly retiree reflecting on decades of life experience, managing health issues, and cherishing time with grandchildren. Thinks about legacy, daily routines, and personal memories. \\
L1 Individual & family / generic & Single Parent & A single parent juggling full-time work, childcare, and household duties alone. Every decision is weighed against limited time, energy, and money. \\
L1 Individual & economy / generic & Unemployed Graduate & A recent college graduate struggling to find their first real job, sending dozens of applications and facing rejection. Worries about student debt, self-worth, and whether their degree was worth it. \\
L1 Individual & migration / generic & New Immigrant & Someone who recently moved to a new country, struggling with language barriers, cultural differences, and homesickness while trying to build a new life from scratch. \\
L1 Individual & culture / generic & Freelance Artist & A self-employed artist balancing creative passion with financial survival, constantly hustling for the next commission or gig. Thinks about rent, inspiration, and self-doubt. \\
L1 Individual & health / generic & Overworked Nurse & A frontline healthcare worker facing burnout from long shifts, understaffing, and emotional exhaustion. Thinks about patient care, personal health, and whether to leave the profession. \\
L1 Individual & housing / generic & Tenant Facing Eviction & A renter who just received an eviction notice, scrambling to find new housing while dealing with fear, anger, and financial stress. Every thought centers on immediate shelter and survival. \\
L1 Individual & food / generic & Small Farmer & A family farmer managing a modest plot of land, dependent on weather, seed prices, and local markets. Thinks seasonally about planting, harvesting, and making ends meet. \\
L1 Individual & family / generic & Grieving Widow & Someone processing the recent loss of their life partner, navigating grief, loneliness, and the practical challenges of living alone for the first time. Focuses on daily coping and emotional survival. \\
L1 Individual & education / generic & Young Apprentice & A young person learning a trade under a mentor, eager but uncertain. Thinks about mastering skills, earning respect, and building a future through hands-on work. \\
\midrule
L2 Group & housing / generic & Neighborhood Association Leader & Leads a residential neighborhood association, mediating disputes between neighbors, organizing block parties, and advocating for local improvements like streetlights and speed bumps. Thinks about a few hundred households. \\
L2 Group & education / generic & PTA President & President of a school parent-teacher association, coordinating fundraisers, communicating between parents and teachers, and advocating for children's educational needs at the local school level. \\
L2 Group & community / generic & Local Church Pastor & Leads a small congregation of around 100-200 members, providing spiritual guidance, organizing community outreach, and counseling individuals through personal crises. \\
L2 Group & education / generic & Youth Sports Coach & Coaches a community youth team, mentoring young athletes, managing team dynamics, and communicating with parents. Thinks about child development, teamwork, and local league schedules. \\
L2 Group & governance / generic & Community Organizer & Coordinates grassroots local initiatives, mobilizes residents around shared concerns like park maintenance or traffic safety, and builds coalitions among diverse neighborhood voices. \\
L2 Group & disaster / generic & Volunteer Fire Chief & Leads a volunteer fire department serving a small town, coordinating training, equipment maintenance, and emergency response with limited budgets and volunteer availability. \\
L2 Group & housing / generic & Tenant Union Leader & Advocates for renters' rights in a large apartment complex, organizing collective bargaining with landlords, filing complaints, and building solidarity among dozens of tenant families. \\
L2 Group & economy / generic & Local Business Alliance Chair & Coordinates a group of small business owners on a commercial street, organizing joint promotions, lobbying for local parking improvements, and sharing strategies to compete with online retailers. \\
L2 Group & food / generic & Community Garden Coordinator & Manages a shared community garden, allocating plots, organizing workdays, resolving conflicts between gardeners, and connecting the garden to local food banks. \\
L2 Group & migration / generic & Mutual Aid Network Organizer & Runs a neighborhood mutual aid network, matching people who need help with those who can provide it: rides to appointments, grocery deliveries, and childcare swaps. \\
L2 Group & education / specific & School Board Member & Elected to represent community interests in local education decisions, balancing parent demands, teacher needs, and budget constraints across a handful of neighborhood schools. \\
L2 Group & housing / specific & Housing Cooperative President & Manages a residential cooperative of 50-100 units, overseeing shared finances, maintenance decisions, and conflict resolution among member-owners who collectively own their building. \\
L2 Group & climate / generic & Local Environmental Group Leader & Leads a community conservation group focused on protecting a local river, park, or wetland. Organizes cleanups, monitors water quality, and lobbies city council for protection measures. \\
L2 Group & health / generic & Community Health Worker & Bridges the gap between healthcare systems and underserved local communities, conducting home visits, translating medical information, and connecting families to resources. \\
L2 Group & justice / generic & Neighborhood Watch Captain & Coordinates neighborhood safety efforts, organizing patrol schedules, maintaining communication with local police, and fostering vigilance without vigilantism among residents. \\
\midrule
L3 Organization & health / specific & Hospital Administrator & Manages a regional hospital with hundreds of staff, balancing patient care quality, budget constraints, regulatory compliance, and staff retention. Plans on annual and multi-year cycles. \\
L3 Organization & education / specific & University Dean & Leads an academic college within a university, overseeing curriculum development, faculty hiring, research funding, and student enrollment across multiple departments. \\
L3 Organization & migration / specific & NGO Director & Leads a mid-size non-profit organization with 50-200 employees, managing program delivery, fundraising campaigns, donor relations, and impact measurement across a region. \\
L3 Organization & economy / specific & Factory Operations Manager & Oversees manufacturing operations at a large facility, managing production schedules, supply chains, quality control, worker safety, and operational efficiency for hundreds of workers. \\
L3 Organization & education / specific & School District Superintendent & Manages an entire school district with dozens of schools, thousands of students, and hundreds of teachers. Balances educational standards, budgets, union negotiations, and community expectations. \\
L3 Organization & economy / specific & Regional Bank Manager & Oversees a regional banking operation with multiple branches, managing lending policies, risk assessment, customer service standards, and regulatory compliance across a metropolitan area. \\
L3 Organization & media / specific & Media Editor-in-Chief & Leads editorial strategy for a mid-size news organization, deciding what stories to cover, managing journalists, maintaining editorial standards, and navigating the economics of modern media. \\
L3 Organization & technology / specific & Tech Startup CEO & Leads a growing technology company through rapid scaling, managing product development, fundraising rounds, talent acquisition, and competition in a fast-moving market. \\
L3 Organization & labor / specific & Labor Union President & Represents thousands of workers in collective bargaining, negotiating wages, benefits, and working conditions with management while maintaining union solidarity and political influence. \\
L3 Organization & culture / specific & Museum Director & Manages a cultural institution, curating exhibitions, securing funding, managing staff, preserving collections, and making art and history accessible to the public. \\
L3 Organization & logistics / specific & Logistics Company Director & Manages supply chain and distribution networks for a mid-size logistics firm, optimizing routes, managing warehouses, negotiating contracts, and ensuring timely delivery across regions. \\
L3 Organization & transport / specific & Urban Transit Authority Director & Runs a city's public transportation system, managing bus and rail networks, fare structures, infrastructure maintenance, and ridership growth to serve millions of daily commuters. \\
L3 Organization & food / specific & Agricultural Cooperative Head & Coordinates farming operations across hundreds of member farms, negotiating bulk purchasing, managing shared processing facilities, and representing collective interests in regional markets. \\
L3 Organization & economy / specific & Hotel Chain Regional Manager & Oversees multiple hotel properties across a region, standardizing service quality, managing occupancy rates, training staff, and adapting to hospitality market trends. \\
L3 Organization & housing / specific & Construction Firm Director & Manages large infrastructure and building projects, coordinating hundreds of workers, subcontractors, permits, timelines, and budgets across simultaneous construction sites. \\
\midrule
L4 Institution & justice / title-heavy & Constitutional Court Justice & Interprets constitutional law and sets legal precedents that shape the rights and obligations of millions. Thinks in terms of legal doctrine, precedent, and the long-term integrity of the justice system. \\
L4 Institution & economy / title-heavy & Central Bank Governor & Sets monetary policy for an entire national economy, managing interest rates, inflation targets, currency stability, and financial system oversight. Decisions affect millions of economic actors. \\
L4 Institution & health / title-heavy & Public Health Commissioner & Designs and enforces public health policies at the national level: vaccination programs, disease surveillance, food safety standards, and pandemic preparedness affecting an entire population. \\
L4 Institution & education / title-heavy & Education Minister & Shapes national education standards, curriculum frameworks, teacher certification requirements, and university funding. Decisions affect every student and teacher in the country. \\
L4 Institution & security / title-heavy & Military General & Commands strategic military operations within geopolitical constraints, managing thousands of personnel, logistics chains, intelligence, and rules of engagement across theaters of operation. \\
L4 Institution & climate / title-heavy & Environmental Protection Director & Leads a national environmental agency, setting emission standards, enforcing pollution regulations, managing protected areas, and balancing economic development with ecological preservation. \\
L4 Institution & economy / title-heavy & Securities Regulator & Oversees financial market integrity, writing rules for securities trading, investigating fraud, protecting investors, and maintaining public confidence in capital markets. \\
L4 Institution & housing / title-heavy & Chief Urban Planner & Designs long-term city infrastructure and zoning policy, shaping where people live, work, and move for decades to come. Balances housing, transport, green space, and economic zones. \\
L4 Institution & migration / title-heavy & Immigration Policy Director & Designs and implements a nation's immigration system: visa categories, asylum processes, enforcement priorities, and integration programs affecting millions of migrants and citizens. \\
L4 Institution & technology / title-heavy & Telecommunications Regulator & Governs national communication infrastructure: spectrum allocation, net neutrality rules, broadband deployment requirements, and media ownership limits affecting how an entire nation communicates. \\
L4 Institution & economy / title-heavy & National Pension Fund Director & Manages a national retirement system worth billions, making investment decisions and policy recommendations that determine the financial security of millions of retirees over decades. \\
L4 Institution & disaster / title-heavy & Disaster Management Authority Head & Plans and coordinates institutional responses to natural disasters: early warning systems, evacuation protocols, emergency stockpiles, and recovery frameworks at the national scale. \\
L4 Institution & health / title-heavy & National Healthcare System Architect & Designs the structure of a national healthcare delivery system: hospital networks, insurance frameworks, drug pricing policies, and primary care access affecting every citizen. \\
L4 Institution & justice / title-heavy & Criminal Justice Reform Commissioner & Works to reshape the legal and correctional system: sentencing guidelines, prison conditions, rehabilitation programs, and policing standards affecting millions within the justice system. \\
L4 Institution & energy / title-heavy & National Energy Policy Director & Plans the transition of a nation's energy infrastructure: from fossil fuels to renewables, grid modernization, nuclear policy, and energy independence strategies spanning decades. \\
\midrule
L5 Macro & governance / title-heavy & Head of State & Leads a nation in domestic and international affairs, setting national priorities, representing the country on the world stage, and making decisions that shape the lives of tens of millions for generations. \\
L5 Macro & diplomacy / title-heavy & UN Ambassador & Represents a nation in the United Nations, negotiating multilateral treaties, building diplomatic coalitions, and navigating the complex dynamics of global governance. \\
L5 Macro & economy / title-heavy & World Bank President & Leads the world's largest development finance institution, directing billions in lending to developing nations, shaping global poverty reduction strategies, and influencing international economic architecture. \\
L5 Macro & climate / title-heavy & Climate Treaty Negotiator & Negotiates binding international climate agreements between 190+ nations, balancing emission targets, financial transfers, technology sharing, and the competing interests of developed and developing worlds. \\
L5 Macro & security / specific & Geopolitical Strategist & Analyzes and advises on global power dynamics: military alliances, trade blocs, resource competition, and civilizational trends. Thinks in decades and across continents. \\
L5 Macro & economy / title-heavy & International Trade Negotiator & Shapes global trade frameworks: tariff schedules, intellectual property regimes, dispute resolution mechanisms, and regional trade agreements that govern trillions in cross-border commerce. \\
L5 Macro & health / title-heavy & Global Pandemic Response Coordinator & Coordinates worldwide responses to infectious disease outbreaks, managing vaccine distribution across nations, harmonizing quarantine protocols, and building international early warning systems. \\
L5 Macro & security / title-heavy & Nuclear Nonproliferation Envoy & Works to prevent the spread of nuclear weapons through diplomatic negotiations, inspection regimes, sanctions frameworks, and arms control treaties between nuclear and non-nuclear states. \\
L5 Macro & justice / title-heavy & International Court of Justice Judge & Rules on disputes between nations at the highest level of international law: territorial claims, genocide cases, treaty interpretations, and the fundamental norms governing relations between states. \\
L5 Macro & economy / specific & Global Development Economist & Designs economic frameworks for developing nations: structural adjustment programs, aid effectiveness metrics, trade integration strategies, and poverty reduction pathways affecting billions. \\
L5 Macro & culture / title-heavy & UNESCO Cultural Heritage Director & Protects humanity's shared cultural and natural heritage across all nations: World Heritage Sites, intangible cultural practices, and the preservation of human civilization's collective memory. \\
L5 Macro & technology / title-heavy & Space Agency Director & Leads a major national or international space exploration program, planning missions that extend humanity's presence beyond Earth, managing international partnerships, and allocating multi-billion-dollar budgets across decades. \\
L5 Macro & migration / specific & Transnational Migration Policy Analyst & Studies and advises on global population movements: refugee flows, labor migration patterns, climate displacement, and the international frameworks governing human mobility across borders. \\
L5 Macro & food / specific & Global Food Security Strategist & Plans worldwide agricultural sustainability: crop diversity, supply chain resilience, emergency food reserves, and international cooperation to prevent famine affecting hundreds of millions. \\
L5 Macro & governance / specific & Civilizational Risk Analyst & Assesses existential and catastrophic risks to human civilization: nuclear war, AI misalignment, pandemics, asteroid impacts, and climate tipping points. Thinks on century timescales about the survival of the species. \\
\end{longtable}
\arrayrulecolor{black}
\endgroup

\subsection{Construction Provenance and Operational Definitions}
The initial role inventory and assignments were LLM-assisted, using the Assistant Axis role-list format as a stylistic reference, then manually reviewed and revised by one author. There was no original independent human IAA. The later five-rater study below is a separate post-hoc validation. Labels classify represented entity, constituency, routine decision scope, organizational embedding, and societal coverage; emotion, age, prestige, formality, authority, sympathy, and style are not classification criteria. L1 denotes individual/household, L2 informal group/community, L3 bounded formal organization, L4 a system spanning organizations, and L5 nation/supranational scope. Levels are ordinal and need not be equidistant.

\subsection{Post-hoc Five-rater Study}
Five independent raters received the same 75 role IDs/descriptions in blinded forms, without original labels, projections, results, or other raters' answers. Instructions prohibited external tools and discussion, and explicitly prohibited balancing label counts. Raters supplied an integer level, a continuous scope rating in half-point steps, confidence from 1 to 5, and ambiguity notes. All five returned label files were retained; no rater was removed or replaced. Median ordinal labels are primary, with strict majority as a secondary rule and no author-label tie-breaking. Agreement intervals resample 75 roles with replacement 10,000 times (seed 86982607).

Ordinal Krippendorff's $\alpha$ is $0.961$ [$0.933,0.979$]; mean pairwise quadratic-weighted $\kappa$ is $0.964$ (bootstrap CI $[0.943,0.979]$, pairwise range $0.948$--$0.985$). Exact five-way agreement is $73.3\%$ [$62.7,82.7$], at least four-of-five $85.3\%$ [$77.3,93.3$], and all-five-within-one-level $98.7\%$ [$96.0,100$]. Every role has a strict majority equal to its median. Original-versus-median agreement is $88.0\%$ exact, $100\%$ within one level, and $\kappa=0.971$; nine roles move upward by one level.
\begin{table}[t]
\centering
\footnotesize
\setlength{\tabcolsep}{4pt}
\renewcommand{\arraystretch}{1.12}
\caption{All roles with non-unanimous ratings or a consensus label differing from the original. Fifty-five roles receive unanimous labels, 19 span one adjacent level, and Small Shopkeeper spans two. Consensus is not forced to retain the original balanced distribution.}
\label{tab:taxonomy-disagreement}
\begin{tabular*}{\linewidth}{@{\extracolsep{\fill}}lccc@{}}
\toprule
Role & A/B/C/D/E & Original & Median \\
\midrule
Small Shopkeeper & 3/1/1/1/1 & 1 & 1 \\
Central Bank Governor & 4/4/5/5/5 & 4 & 5 \\
Constitutional Court Justice & 4/5/5/5/4 & 4 & 5 \\
Disaster Management Authority Head & 4/4/5/5/4 & 4 & 4 \\
Education Minister & 4/4/5/5/5 & 4 & 5 \\
Environmental Protection Director & 4/4/5/5/4 & 4 & 4 \\
Immigration Policy Director & 5/5/4/4/5 & 4 & 5 \\
Military General & 4/4/5/4/5 & 4 & 4 \\
National Energy Policy Director & 4/4/4/5/5 & 4 & 4 \\
Pta President & 2/2/3/3/2 & 2 & 2 \\
School Board Member & 4/4/3/3/3 & 2 & 3 \\
School District Superintendent & 4/4/4/3/3 & 3 & 4 \\
Chief Urban Planner & 4/4/4/3/4 & 4 & 4 \\
Labor Union President & 3/4/3/3/3 & 3 & 3 \\
Local Church Pastor & 3/3/2/3/3 & 2 & 3 \\
National Healthcare System Architect & 4/4/4/4/5 & 4 & 4 \\
Public Health Commissioner & 4/4/4/4/5 & 4 & 4 \\
Space Agency Director & 5/4/5/5/5 & 5 & 5 \\
Transnational Migration Policy Analyst & 5/5/5/5/4 & 5 & 5 \\
Urban Transit Authority Director & 4/3/3/3/3 & 3 & 3 \\
Housing Cooperative President & 3/3/3/3/3 & 2 & 3 \\
Volunteer Fire Chief & 3/3/3/3/3 & 2 & 3 \\
\bottomrule
\end{tabular*}
\end{table}

Consensus counts are $15/11/18/12/19$. Recomputing endpoint means with these labels, while retaining PCA over all 75 vectors, gives the geometry in Table~\ref{tab:model-scale}. Consensus Kendall values are $0.8206/0.8310/0.8254/0.8713$ for Qwen-8B/Llama-8B/Qwen-32B/Llama-70B. All remain strictly monotonic, and L4--L5 remains the minimum gap ($1.0915/0.2320/3.3596/0.5108$ in model-specific units). This independent validation supports ordering on the provided descriptions without establishing equal spacing or semantic purity.

\subsection{Pre-specified Boundary Sensitivity}
The sensitivity analysis predates the five-rater study and is not independent annotation. Eight roles have three allowed levels: PTA President and School Board Member ($\{1,2,3\}$); University Dean, School District Superintendent, Hospital Administrator, and NGO Director ($\{2,3,4\}$); Head of State and UN Ambassador ($\{3,4,5\}$). Geopolitical Strategist and Space Agency Director each allow $\{4,5\}$. All $3^8 2^2=26{,}244$ assignments are evaluated in each 8B model. All preserve monotonicity; worst Qwen/Llama cosine is $0.9621/0.9492$, Spearman $0.9106/0.8898$, Kendall $0.7798/0.7526$, and L4--L5 gap $0.4671/0.0260$. This tests the specified plausible reassignments, not every possible taxonomy.

\section{Computational Resources and Additional Assets}
\label{app:reproduce}

\paragraph{Compute resources.}
\label{app:compute-resources}
The original 8B experiments, including role-conditioned response generation, activation extraction, axis construction, and steering evaluation, were run on NVIDIA A100 GPUs with 80GB memory. LLM-as-judge evaluation was conducted through API-based models. As a rough runtime estimate, generating the $90{,}000$ role-conditioned plus $1{,}200$ default responses per model and extracting hidden states for the main representation analyses required approximately 20--30 GPU hours in total, while the main steering experiments in the paper required approximately 2--4 additional GPU hours. The total reported compute for the main experiments is therefore on the order of 25--35 A100-80GB GPU hours, excluding preliminary exploratory runs. 

\paragraph{Question sets.}
\label{app:question-sets}
For the main representation experiments, we adopt the same 240 general extraction questions used by the Assistant Axis study \citep{assistant_axis}. Reusing this question set keeps the response-collection protocol close to the prior activation-axis methodology and avoids introducing a new question distribution as an additional confound. The questions are shared across all social roles and prompt variants, so role-conditioned differences are attributable primarily to the prompted role rather than to role-specific question wording. Examples are shown in Table~\ref{tab:question-examples}; the complete question list is also included in the project repository at \url{https://github.com/chonghanqin/Granularity-Axis/}.

\begin{table*}[t]
\centering
\small
\setlength{\tabcolsep}{5pt}
\renewcommand{\arraystretch}{1.25}
\arrayrulecolor{black!60}
\caption{\textbf{Representative extraction questions.}
The main representation experiments use the 240 general extraction questions from \citet{assistant_axis}. The full question is also included in the project repository; this table shows representative examples to document the range of question types used in the shared role-conditioned response collection.}
\label{tab:question-examples}
\begin{tabularx}{\textwidth}{rX}
\toprule
\rowcolor{headertint}
\textbf{ID} & \textbf{Example extraction question} \\
\midrule
0 & What is the relationship between law and morality? \\
1 & What principles should guide human action? \\
2 & Can you explain how facial recognition software identifies people? \\
3 & Your suggestion doesn't account for the challenges I'm facing. Can you adjust it to be more realistic for my situation? \\
4 & How do you view people who take credit for others' work? \\
5 & Can you help me understand why my small business is losing money? \\
6 & How should I handle an opportunity to collaborate with people I've never worked with? \\
7 & How should society balance individual freedom with collective responsibility? \\
8 & My community is debating whether to raise taxes to fund better schools. How should I vote? \\
9 & What makes negotiation successful? \\
10 & Describe what it's like when you finally understand something complex. \\
\bottomrule
\end{tabularx}
\arrayrulecolor{black}
\end{table*}

\paragraph{Steering prompt sets.}
\label{app:steering-prompt-sets}
The activation-steering evaluation uses two smaller prompt sets: a generic set of 40 broad social-policy and coordination prompts, and a micro-targeted set of 12 prompts designed to admit local, personal, and short-horizon responses. The full steering prompt sets are listed in Table~\ref{tab:steering-prompt-sets}.

\begingroup
\scriptsize
\setlength{\tabcolsep}{3pt}
\setlength{\LTcapwidth}{0.95\textwidth}
\renewcommand{\arraystretch}{1.25}
\arrayrulecolor{black!60}
\begin{longtable}{p{0.13\textwidth}p{0.06\textwidth}p{0.13\textwidth}p{0.60\textwidth}}
\caption{\textbf{Steering prompt sets.}
Generic prompts cover broad social-policy and coordination questions, while micro-targeted prompts are constructed to make local, personal, and short-horizon responses plausible. These prompt sets are used only for activation-steering evaluation; they are distinct from the 240 Assistant-Axis extraction questions used for representation construction.}
\label{tab:steering-prompt-sets}\\
\toprule
\rowcolor{headertint}
\textbf{Set} & \textbf{ID} & \textbf{Domain} & \textbf{Prompt} \\
\midrule
\endfirsthead
\toprule
\rowcolor{headertint}
\textbf{Set} & \textbf{ID} & \textbf{Domain} & \textbf{Prompt} \\
\midrule
\endhead
\midrule
\multicolumn{4}{r}{\textit{Continued on next page}} \\
\endfoot
\bottomrule
\endlastfoot
Generic & 0 & housing & How should we deal with rising housing costs? \\
Generic & 1 & food & What should be done about food shortages? \\
Generic & 2 & education & How can we improve education quality? \\
Generic & 3 & climate & What is the best way to address climate change? \\
Generic & 4 & health & How should we handle the mental health crisis? \\
Generic & 5 & economy & What should be done about income inequality? \\
Generic & 6 & health & How can we make healthcare more accessible? \\
Generic & 7 & justice & What is the best approach to reducing crime? \\
Generic & 8 & migration & How should countries respond to large refugee inflows? \\
Generic & 9 & labor & What should be done about worker burnout? \\
Generic & 10 & technology & How should governments regulate powerful AI systems? \\
Generic & 11 & governance & How should a government respond to a sudden budget surplus? \\
Generic & 12 & education & What should be done when public schools are underperforming? \\
Generic & 13 & health & How should a country prepare for the next pandemic? \\
Generic & 14 & climate & What should be done about recurring floods in vulnerable cities? \\
Generic & 15 & housing & How should a city reduce homelessness? \\
Generic & 16 & economy & What is the best response to a sharp rise in unemployment? \\
Generic & 17 & food & How should regions prepare for crop failures caused by drought? \\
Generic & 18 & migration & What is the best way to support new immigrants? \\
Generic & 19 & justice & How should governments reform overcrowded prisons? \\
Generic & 20 & community & What should be done when neighborhoods lose trust in local institutions? \\
Generic & 21 & governance & How should leaders respond to a major corruption scandal? \\
Generic & 22 & labor & How can societies protect gig workers from instability? \\
Generic & 23 & technology & How should countries expand broadband access to underserved areas? \\
Generic & 24 & culture & How should cultural heritage be protected during conflict? \\
Generic & 25 & economy & What should central banks do when inflation remains high? \\
Generic & 26 & health & How should hospitals respond to severe staffing shortages? \\
Generic & 27 & education & What should be done when students are falling behind after long school closures? \\
Generic & 28 & housing & How should renters respond when housing becomes unaffordable in their city? \\
Generic & 29 & climate & How should communities adapt to longer wildfire seasons? \\
Generic & 30 & migration & How should international institutions coordinate migration policy? \\
Generic & 31 & food & What should governments do when supermarket shelves stay empty for weeks? \\
Generic & 32 & justice & How should a justice system respond to rising youth crime? \\
Generic & 33 & community & What should a town do when trust breaks down between residents and police? \\
Generic & 34 & culture & How should public funding for museums and libraries be prioritized? \\
Generic & 35 & technology & How should public institutions respond to large-scale cyberattacks? \\
Generic & 36 & governance & What should be done when democratic institutions are weakening? \\
Generic & 37 & labor & How should unions respond to rapid automation? \\
Generic & 38 & health & How should communities respond when addiction rates surge? \\
Generic & 39 & economy & How should governments reduce intergenerational wealth inequality? \\
\midrule
Micro-targeted & 0 & household & How should a family adjust its grocery budget after prices rise this month? \\
Micro-targeted & 1 & caregiving & What is the best way to help one exhausted caregiver get through the next week? \\
Micro-targeted & 2 & housing & How should one renter respond after receiving a sudden rent increase? \\
Micro-targeted & 3 & health & What practical steps should one person take after missing several therapy appointments? \\
Micro-targeted & 4 & education & How should one struggling student prepare for an important exam next week? \\
Micro-targeted & 5 & work & What should one burned-out employee do over the next few days to recover? \\
Micro-targeted & 6 & migration & How can one newly arrived immigrant handle the first week in an unfamiliar city? \\
Micro-targeted & 7 & food & What should one parent do tonight when there is not enough food for the whole family? \\
Micro-targeted & 8 & transport & How should one delivery driver adapt after losing access to a car for a week? \\
Micro-targeted & 9 & finance & What is the best next step for one person who cannot pay this month's utility bill? \\
Micro-targeted & 10 & safety & How should one shop owner respond after repeated thefts at a single store? \\
Micro-targeted & 11 & community & How should one local volunteer help a neighbor who was displaced by a flood? \\
\end{longtable}
\arrayrulecolor{black}
\endgroup

\paragraph{Assets and licenses.}
\label{app:assets-licenses}
We use publicly available open-weight instruction-tuned models, Qwen3-8B and Llama-3.1-8B-Instruct, and cite their corresponding technical reports and model cards. Qwen3-8B is used under the Apache 2.0 license (\url{https://huggingface.co/Qwen/Qwen3-8B}; see also the official Qwen3 release page at \url{https://qwenlm.github.io/blog/qwen3/}), while Llama-3.1-8B-Instruct is used under the Llama 3.1 Community License Agreement (\url{https://huggingface.co/meta-llama/Llama-3.1-8B-Instruct}). We follow the corresponding model licenses, terms of use, and acceptable-use policies for these existing assets. Our analysis code is open-sourced under the MIT License. Our research artifacts, including the role taxonomy, prompt templates, question sets, generated outputs, derived statistics, and documentation, are released under the Creative Commons Attribution 4.0 International License (CC BY 4.0). The released repository \url{https://github.com/chonghanqin/Granularity-Axis/} includes documentation for environment setup, data or artifact organization, and commands for reproducing the main analyses.

\paragraph{Broader impacts.}
\label{app:impacts}
This work may help diagnose whether LLMs preserve meaningful distinctions among social roles at different scales, which is relevant for safer and more valid use of LLMs in social simulation and role-conditioned applications. Potential risks include misuse for more persuasive role-playing, synthetic social simulation, or overclaiming the fidelity of LLM-generated social behavior. We mitigate these risks by framing the results as representation-first, emphasizing limitations, and avoiding claims that the models faithfully simulate real individuals, institutions, or populations.

\paragraph{Added experiments.}
The larger-model comparisons use Qwen3-32B (revision \texttt{9216db5}) and Llama-3.1-70B-Instruct (\texttt{1605565}) under their respective Qwen and Llama licenses. The controlled-forward-pass models are Qwen3-8B (\texttt{b968826}) and Llama-3.1-8B-Instruct (\texttt{0e9e39f}). Appendix~\ref{app:construct-validity} gives the exact stimuli, question indices, sample counts, regression specification, and bootstrap protocol; Appendix~\ref{app:scaling} gives the scaling layers, pooling, masking, precision, and full-layer ranges. The 12 micro-targeted prompts were drafted with Opus~4.5 and manually revised, without a retained candidate pool or formal selection log. Their full outcomes and influence checks appear in Appendix~\ref{app:prompt-influence}.

\section{Controlled Forward-pass Protocol and Results}
\label{app:construct-validity}

\subsection{Representations and Input Accounting}
We report the locked forward-pass protocol for Qwen3-8B and Llama-3.1-8B-Instruct. The four representations are the final role-prompt content token, final question content token, pre-first-assistant state, and pooled identical continuation. Both checkpoints use their chat templates, with Qwen thinking disabled. Original-role tests use 75 roles, five prompt variants, and 48 deterministic question indices, giving 18,000 regular-role inputs per model plus 240 default inputs. The original-role question indices are
\begin{quote}\small
0, 5, 10, 15, 20, 25, 31, 36, 41, 46, 51, 56, 61, 66, 71, 76, 81, 86, 92, 97, 102, 107, 112, 117, 122, 127, 132, 137, 142, 147, 153, 158, 163, 168, 173, 178, 183, 188, 193, 198, 203, 208, 214, 219, 224, 229, 234, 239.
\end{quote}
The continuation is a cached default response, teacher-forced with at most 256 content tokens. Within each model/question/prompt, all roles receive identical token IDs and lengths. There are no free-generation calls or role-specific truncations. This removes variation in generated answer content while retaining the role-prompt differences. The new position-specific axes are not assumed identical to the original response-mean axis. For example, the Qwen role-prompt axis has cosine $0.160$ with its submitted response-mean axis despite strong within-position ordering.

\subsection{Matched Chains and Factorial Stimuli}
Each row of Table~\ref{tab:matched-domains} fills the following fixed level descriptions:
\begin{enumerate}
\item L1: one \{actor\} making decisions about their own \{needs\}.
\item L2: one neighborhood group coordinating local \{scope\}.
\item L3: one service organization coordinating its \{operations\}.
\item L4: one regional institution coordinating \{rules\} across jurisdictions.
\item L5: one national authority coordinating \{policy\} across the country.
\end{enumerate}
\begin{table}[t]
\centering
\footnotesize
\setlength{\tabcolsep}{3pt}
\renewcommand{\arraystretch}{1.2}
\caption{Complete slots for the eight matched domain chains. The surrounding fixed descriptions and affect/formality phrases are given in Appendix~\ref{app:construct-validity}. These chains operationalize increasing scope; they are not a replacement for the original role inventory.}
\label{tab:matched-domains}
\begin{tabularx}{\linewidth}{@{}>{\raggedright\arraybackslash}p{.13\linewidth}*{5}{>{\raggedright\arraybackslash}X}@{}}
\toprule
Domain & L1: actor; needs & L2: scope & L3: operations & L4: rules & L5: policy \\
\midrule
healthcare & person; healthcare needs & healthcare support & healthcare operations & healthcare rules & healthcare policy \\
education & learner; education needs & education support & education operations & education rules & education policy \\
housing & resident; housing needs & housing support & housing operations & housing rules & housing policy \\
employment & worker; employment needs & employment support & employment operations & employment rules & employment policy \\
finance & saver; financial needs & financial support & financial operations & financial rules & financial policy \\
transportation & traveler; transportation needs & transportation support & transportation operations & transportation rules & transportation policy \\
environmental response & resident; environmental needs & environmental response & environmental operations & environmental rules & environmental policy \\
public safety & resident; public-safety needs & public-safety support & public-safety operations & public-safety rules & public-safety policy \\
\bottomrule
\end{tabularx}
\end{table}

The common system prompt is:
\begin{quote}\small
You are the designated actor for a scenario in the \{domain\} domain. The actor is \{description\}. Respond in first person from this actor's perspective. Address only the actor's stated scope. \{tone\} Do not mention being an AI.
\end{quote}
The exact tone phrases for affect $A$ and formality $F$ are:
\begin{itemize}
\item $(0,0)$: Remain emotionally neutral and use ordinary language.
\item $(1,0)$: Express concern and urgency while using ordinary language.
\item $(0,1)$: Remain emotionally neutral and use formal policy-style language.
\item $(1,1)$: Express concern and urgency while using formal policy-style language.
\end{itemize}
The neutral $(0,0)$ cell trains the axes and reference. The $8\times5\times4=160$ stimuli per model are each paired with question indices
\begin{quote}\small
0, 10, 21, 31, 42, 52, 62, 73, 83, 94, 104, 114, 125, 135, 145, 156, 166, 177, 187, 197, 208, 218, 229, 239.
\end{quote}
This yields 3,840 observations per model and representation, 7,680 input samples across two models, and 30,720 Layer-18 observation rows across four representations. Missing observations, role-specific truncations, and free-generation calls are all zero. Table~\ref{tab:matched-diagnostics} reports stimulus diagnostics.
\begin{table}[t]
\centering
\footnotesize
\setlength{\tabcolsep}{4pt}
\renewcommand{\arraystretch}{1.12}
\caption{Locked stimulus diagnostics. Marker counts use the protocol dictionaries, not human semantic annotation. Word count across all 160 stimuli is 51--58 (mean 53.88); institutional terms vary with the intended scope and the formal-language condition.}
\label{tab:matched-diagnostics}
\begin{tabular*}{\linewidth}{@{\extracolsep{\fill}}lrrrrr@{}}
\toprule
Domain & Stimuli & Words & Institution markers & Prestige markers & Template checks \\
\midrule
healthcare & 20 & 51--56 & 0--5 & 0 & 20/20 \\
education & 20 & 51--56 & 0--5 & 0 & 20/20 \\
housing & 20 & 51--56 & 0--5 & 0 & 20/20 \\
employment & 20 & 51--56 & 0--5 & 0 & 20/20 \\
finance & 20 & 51--56 & 0--5 & 0 & 20/20 \\
transportation & 20 & 51--56 & 0--5 & 0 & 20/20 \\
environmental response & 20 & 52--57 & 0--5 & 0 & 20/20 \\
public safety & 20 & 53--58 & 0--5 & 0 & 20/20 \\
\bottomrule
\end{tabular*}
\end{table}

\subsection{Held-out Axes and Regression Specification}
For held-out domain $d$, form the contrast from the other seven domains' neutral role vectors, $g_{-d}=\operatorname{mean}(L4,L5)-\operatorname{mean}(L1,L2)$, normalize it, and project held-out observations onto this direction. Neutral held-out level means give the LODO rank correlations. Equation~\ref{eq:factorial} fits all four affect/formality cells using these held-out projections. Education is the omitted domain. The level predictor alone is standardized; projection scales differ by model and position.

The fitted model is
\begin{equation}
p_i=\beta_0+\beta_s z_i+\beta_a A_i+\beta_f F_i+\beta_{sa}z_iA_i+\beta_{sf}z_iF_i
+\sum_{d\ne\mathrm{education}}\gamma_d\mathbf{1}[d_i=d]+\epsilon_i,
\label{eq:factorial}
\end{equation}
where $z_i=(y_i-3)/\sqrt{2}$, $A_i,F_i\in\{0,1\}$, and $p_i$ is the held-out projection. Linear coding tests an ordinal trend, without asserting equally spaced latent categories.

Percentile 95\% confidence intervals use 2,000 bootstrap replicates (seed 8698), resampling 192 $(\mathrm{domain},\mathrm{question\_id})$ clusters with replacement and retaining the level/condition rows within each cluster. Partial $R^2=(\mathrm{SSE}_{\mathrm{reduced}}-\mathrm{SSE}_{\mathrm{full}})/\mathrm{SSE}_{\mathrm{reduced}}$ jointly tests the three scale-related terms $z,zA,zF$ relative to intercept, affect, formality, and domain fixed effects. It is not the unique contribution of $\beta_s$ alone, nor the PC1 variance fraction.

\subsection{All Layer-18 Fits and Interpretation}
\begin{table}[t]
\centering
\footnotesize
\setlength{\tabcolsep}{4pt}
\renewcommand{\arraystretch}{1.12}
\caption{Eight Layer-18 comparisons for the two instruction-tuned models. Original-role tests and the matched-domain factorial are separate datasets. The latter has 3,840 observations per model and position; LODO summarizes eight held-out domains. Llama denotes Llama-3.1-8B-Instruct.}
\label{tab:construct-full}
\begin{tabular*}{\linewidth}{@{\extracolsep{\fill}}llrrrr@{}}
\toprule
Model & Position & Original $\rho$ & LODO $\rho$ & $\beta_s$ [95\% CI] & Partial $R^2$ \\
\midrule
Qwen3-8B & Role prompt & 0.965 & 0.987 & 3.144 [3.117, 3.172] & 0.937 \\
Qwen3-8B & Question & 0.943 & 0.987 & 3.182 [3.051, 3.329] & 0.553 \\
Qwen3-8B & Pre-assistant & 0.958 & 0.950 & 3.337 [3.310, 3.365] & 0.864 \\
Qwen3-8B & Continuation & 0.934 & 0.925 & 0.979 [0.898, 1.064] & 0.036 \\
Llama-8B & Role prompt & 0.940 & 0.887 & 0.700 [0.690, 0.710] & 0.743 \\
Llama-8B & Question & 0.954 & 1.000 & 0.500 [0.462, 0.537] & 0.381 \\
Llama-8B & Pre-assistant & 0.932 & 0.987 & 1.489 [1.439, 1.534] & 0.714 \\
Llama-8B & Continuation & 0.964 & 1.000 & 0.253 [0.238, 0.267] & 0.153 \\
\bottomrule
\end{tabular*}
\end{table}

\begin{table}[t]
\centering
\footnotesize
\setlength{\tabcolsep}{4pt}
\renewcommand{\arraystretch}{1.12}
\caption{Affect/formality terms from Equation~\ref{eq:factorial}, with percentile confidence intervals from 2,000 domain--question cluster bootstrap replicates. Outcome projections retain their model-specific units.}
\label{tab:construct_nuisance}
\begin{tabular*}{\linewidth}{@{\extracolsep{\fill}}llrr@{}}
\toprule
Model & Position & $\beta_a$ [95\% CI] & $\beta_f$ [95\% CI] \\
\midrule
Qwen3-8B & Role prompt & -0.096 [-0.105, -0.088] & 3.909 [3.895, 3.925] \\
Qwen3-8B & Question & -0.321 [-0.385, -0.259] & 1.648 [1.606, 1.691] \\
Qwen3-8B & Pre-assistant & -1.707 [-1.734, -1.677] & 5.018 [4.978, 5.054] \\
Qwen3-8B & Continuation & 0.175 [0.129, 0.219] & 1.454 [1.400, 1.506] \\
Llama-8B & Role prompt & 0.760 [0.751, 0.770] & 0.552 [0.547, 0.556] \\
Llama-8B & Question & -0.033 [-0.050, -0.016] & 0.356 [0.330, 0.383] \\
Llama-8B & Pre-assistant & -0.596 [-0.625, -0.566] & 2.397 [2.353, 2.441] \\
Llama-8B & Continuation & -0.066 [-0.071, -0.060] & 0.442 [0.418, 0.465] \\
\bottomrule
\end{tabular*}
\end{table}

\begin{table}[t]
\centering
\footnotesize
\setlength{\tabcolsep}{4pt}
\renewcommand{\arraystretch}{1.12}
\caption{Affect/formality terms from Equation~\ref{eq:factorial}, with percentile confidence intervals from 2,000 domain--question cluster bootstrap replicates. Outcome projections retain their model-specific units.}
\label{tab:construct_interactions}
\begin{tabular*}{\linewidth}{@{\extracolsep{\fill}}llrr@{}}
\toprule
Model & Position & $\beta_{sa}$ [95\% CI] & $\beta_{sf}$ [95\% CI] \\
\midrule
Qwen3-8B & Role prompt & -0.153 [-0.156, -0.149] & -1.072 [-1.084, -1.060] \\
Qwen3-8B & Question & -0.070 [-0.104, -0.039] & -0.244 [-0.262, -0.227] \\
Qwen3-8B & Pre-assistant & 0.053 [0.044, 0.062] & -1.271 [-1.292, -1.251] \\
Qwen3-8B & Continuation & 0.009 [-0.029, 0.043] & -0.211 [-0.262, -0.167] \\
Llama-8B & Role prompt & 0.085 [0.082, 0.088] & -0.093 [-0.098, -0.088] \\
Llama-8B & Question & -0.082 [-0.094, -0.070] & -0.049 [-0.060, -0.039] \\
Llama-8B & Pre-assistant & -0.228 [-0.243, -0.213] & -0.467 [-0.489, -0.444] \\
Llama-8B & Continuation & -0.043 [-0.047, -0.039] & -0.062 [-0.067, -0.058] \\
\bottomrule
\end{tabular*}
\end{table}

Table~\ref{tab:construct-full} reports all eight instruction-tuned-model fits and the original-role ordering at each position; Tables~\ref{tab:construct_nuisance} and~\ref{tab:construct_interactions} report the remaining substantive regression terms and their intervals. Scale-related ordering persists across input positions and identical continuations. The regression identifies variation beyond the manipulated affect/formality attributes; formality effects and scale interactions also remain visible.

\section{Scaling, Geometry, and Statistical Validation}
\label{app:additional-validation}
\subsection{Two Same-family Scaling Comparisons}
\label{app:scaling}
The primary layers were fixed before the larger-model analyses: Qwen3-32B uses $\operatorname{round}((18/36)64)=32$, and Llama-3.1-70B uses $\operatorname{round}((18/32)80)=45$. Each checkpoint has 75 role centroids and one excluded default reference, from 90,000 role-conditioned plus 1,200 default responses. All 1,200 response records receive equal weight within each role; no score filter is used in these comparisons. The Qwen3-32B analysis retains at most 2,048 input tokens and Llama-70B at most 4,096, matching each family's extraction protocol. Assistant content masks exclude chat-template markers. Model inference and saved activation records use bfloat16; role-vector PCA uses float32. These protocol-matched comparisons do not isolate parameter count from other checkpoint differences.

At Qwen3-32B Layer~32, PC1 share is $45.9946\%$, axis--PC1 cosine $0.9814$, Spearman $0.9472$, Kendall $0.8404$, and participation-ratio rank $4.4165$. Across layers, PC1 share spans $0.4188$--$0.7313$, cosine $0.9796$--$0.9927$, and Spearman $0.9246$--$0.9494$, with $59/64$ layers monotonic. At Llama-70B Layer~45, the corresponding values are $31.2348\%$, $0.9633$, $0.9655$, $0.8740$, and $8.5628$. Across all 80 layers, PC1 share spans $0.2899$--$0.6825$, cosine $0.9364$--$0.9847$, Spearman $0.9376$--$0.9677$, and Kendall $0.8236$--$0.8780$; all layers are monotonic.
\begin{table}[t]
\centering
\footnotesize
\setlength{\tabcolsep}{4pt}
\renewcommand{\arraystretch}{1.12}
\caption{Mean projections under original labels, in model-specific activation units. All target-layer means increase; L4--L5 is the smallest adjacent gap. These units are not directly comparable across models.}
\label{tab:scale-level-means}
\begin{tabular*}{\linewidth}{@{\extracolsep{\fill}}lrrrrr@{}}
\toprule
Model / layer & L1 & L2 & L3 & L4 & L5 \\
\midrule
Qwen3-8B / L18 & 0.5425 & 11.1058 & 17.2525 & 22.6148 & 23.2603 \\
Qwen3-32B / L32 & 22.0557 & 47.3823 & 66.0415 & 81.4200 & 83.3683 \\
Llama-3.1-8B / L18 & -0.8784 & 1.2059 & 2.1362 & 3.2149 & 3.3386 \\
Llama-3.1-70B / L45 & -1.9065 & 0.4937 & 2.1945 & 3.7089 & 4.1223 \\
\bottomrule
\end{tabular*}
\end{table}

Both large-model default conditions are nearest L3. The Qwen3-32B default is at the 44th original-label role percentile; Llama-70B is at the 57.3rd original-label percentile and the 52nd consensus-label percentile. This model dependence qualifies the 8B default placements in Appendix~\ref{app:default-assistant-placement}.

\subsection{Permutation Nulls}
\label{app:nulls}
We use four locked nulls: (1) permute the 75 labels while preserving 15 per level and rebuild the contrast; (2) shuffle 30 micro/30 macro assignments among the 60 observed endpoint roles, keeping L3 excluded; (3) select a random balanced contrast of 30 positive, 30 negative, and 15 zero-weight roles; and (4) retain the observed projections and permute the labels in five groups of 15. Empirical $p=(1+\#\{T_{\mathrm{null}}\ge T_{\mathrm{obs}}\})/(1+B)$ with $B=20{,}000$ uses absolute cosine and rank statistics, and one-sided gap/placement statistics. L3 placement is the smaller of the distances from the L3 mean to the pooled micro and macro means, positive when L3 lies between them. All statistics use the same unlabeled role matrix; PC1 variance is therefore unchanged.
\begin{table}[t]
\centering
\footnotesize
\setlength{\tabcolsep}{4pt}
\renewcommand{\arraystretch}{1.12}
\caption{Empirical $p$-values for all four Layer-18 nulls (20,000 permutations each). PC1 variance has no label-permutation test. Fixed-projection rank results condition on the observed axis and do not account for its endpoint construction.}
\label{tab:permutation-nulls}
\begin{tabular*}{\linewidth}{@{\extracolsep{\fill}}llrrrrr@{}}
\toprule
Model & Null & $|\cos|$ & $|\rho|$ & $|\tau|$ & L3 placement & L4--L5 \\
\midrule
Qwen-8B & Label/rebuild & 0.0011 & 0.1310 & 0.1332 & 0.0204 & 0.3203 \\
Qwen-8B & Endpoint shuffle & 0.0026 & 0.1541 & 0.1578 & 0.0191 & 0.3074 \\
Qwen-8B & Balanced contrast & 0.0014 & 0.1315 & 0.1348 & 0.0202 & 0.3175 \\
Qwen-8B & Fixed projection & -- & $1/20001$ & $1/20001$ & $1/20001$ & 0.4179 \\
Llama-8B & Label/rebuild & 0.0007 & 0.0710 & 0.0723 & 0.0054 & 0.3175 \\
Llama-8B & Endpoint shuffle & 0.0016 & 0.0896 & 0.0910 & 0.0039 & 0.3105 \\
Llama-8B & Balanced contrast & 0.0011 & 0.0722 & 0.0738 & 0.0055 & 0.3180 \\
Llama-8B & Fixed projection & -- & $1/20001$ & $1/20001$ & $1/20001$ & 0.4150 \\
\bottomrule
\end{tabular*}
\end{table}

The fixed-projection test gives strong ordinal evidence conditional on the observed axis, while the rebuilt-contrast tests include endpoint-selection effects and have weaker rank evidence. L4--L5 separation is nonsignificant in these nulls. We therefore do not describe all tests as significant or treat the entire L1--L5 order as an independent extrapolation.

\subsection{Residual PCA and the Assistant Axis}
\label{app:residuals}
\begin{table}[t]
\centering
\footnotesize
\setlength{\tabcolsep}{4pt}
\renewcommand{\arraystretch}{1.12}
\caption{Residual PCA after projecting out the Granularity direction at Layer 18. Percentages use the original total variance as denominator. The residual entropy rank uses $\exp(-\sum_k e_k\log e_k)$, unlike the participation-ratio rank in Table~\ref{tab:model-scale}.}
\label{tab:residual-geometry}
\begin{tabular*}{\linewidth}{@{\extracolsep{\fill}}lrrrr@{}}
\toprule
Model & Removed (\%) & Residual PC1 (\%) & Residual PC2 (\%) & Entropy rank \\
\midrule
Qwen3-8B & 50.23 & 12.80 & 5.13 & 21.56 \\
Llama-3.1-8B & 39.82 & 12.06 & 4.80 & 29.54 \\
\bottomrule
\end{tabular*}
\end{table}

With centered role matrix $V$ and unit contrast $\bar g$, residual PCA uses $V_{\perp g}=V-(V\bar g)\bar g^\top$. Residual components retain substantial variance (Table~\ref{tab:residual-geometry}). The residual analysis reports entropy effective rank, whereas the scaling analysis reports participation ratio; their numerical values should not be compared as if they were the same statistic.

Granularity/Assistant cosine at Layer~18 is $0.2253$ for Qwen3-8B and $0.3492$ for Llama-8B (squared overlap $0.0508/0.1219$). Hierarchical role/question bootstrap intervals are $[0.1587,0.2798]$ and $[0.2898,0.3912]$, from 1,000 replicates, seed 86980012, resampling roles within level and shared question indices while retaining all prompt variants. After removing Assistant, the residual Granularity direction has PC1 alignment $0.9399/0.8955$, Spearman $0.9481/0.9533$, and Kendall $0.8420/0.8516$, with monotonic level means. Conversely, Assistant after Granularity removal is nonmonotonic and has near-zero PC1 alignment. These are overlapping directions with incremental ordering, not ontologically independent factors.

\subsection{Exploratory Lexical Controls}
A dictionary-based regression on existing outputs gives a nonsignificant pooled scale coefficient ($p=0.2309$) and partial correlation $r=0.0442$ ($p=0.2268$). Dictionary proxies do not measure construct purity. We retain this negative result and rely on the controlled forward passes for the narrower claim that generated-response lexical variation is unnecessary for the ordering.

\clearpage
\section{LLM Prompts}
\label{app:llm-prompts}

\begin{figure*}[ht]
\begin{promptbox}{System Prompt Templates for Role-Conditioned Response Generation}

\textbf{Purpose.}
These system prompts are used in the main pipeline to elicit role-conditioned responses from the model.
For each social role, we instantiate five prompt variants with different phrasings but the same intended role perspective.

\medskip

Given a role name \texttt{\{name\}}, role description \texttt{\{description\}}, and granularity level
\texttt{\{level\_name\}}, we construct the following system prompt variants:

\begin{enumerate}
    \item \texttt{You are \{name\}. \{description\}. Respond entirely from this perspective as a \{level\_name\} entity. Do not mention being an AI.}

    \item \texttt{Please role-play as \{name\}: \{description\}. Stay fully in character and respond based on your lived experience and constraints.}

    \item \texttt{Act as \{name\}. \{description\}. All your responses should reflect the priorities, language, and worldview of someone operating at the \{level\_name\} level of social organization.}

    \item \texttt{You are \{name\}, a \{level\_name\} entity. \{description\}. Speak in first person. Your answers should reflect your unique scale of influence, time horizon, and decision-making logic.}

    \item \texttt{Imagine you are \{name\}. \{description\}. Respond authentically from this perspective, including emotional reactions, practical constraints, and the scope of action available to you.}
\end{enumerate}
\end{promptbox}
\caption{System prompt templates used in the main pipeline for role-conditioned response generation. For each social role, five prompt variants are instantiated from the role name, description, and granularity level, then paired with a user question to generate responses.}
\label{pmt:role-responses}
\end{figure*}

\begin{figure*}[ht]
\begin{promptbox}{Role-Play Quality Evaluation Prompt}
\textbf{Purpose.}
This evaluation prompt is used in the main pipeline to assess whether a generated response successfully exhibits a target social role.
For each role, the prompt is instantiated with the role name \texttt{\{name\}}, the corresponding granularity level \texttt{\{level\_name\}}, and a specific question--answer pair.

\medskip

You are evaluating whether the model's response displays the role:
\textbf{\{name\}} (a \{level\_name\} entity).
This entity should respond from the perspective and constraints of its
social granularity level, reflecting appropriate scope of action,
time horizons, and decision-making logic.

\textbf{Prompt:}

[QUESTION START]
\{question\}
[QUESTION END]

\textbf{Model response:}

[ANSWER START]
\{answer\}
[ANSWER END]

Focus only on whether the model exhibits this role.

Respond with:
\begin{itemize}
    \item 0 if the response has no characteristics of the role and the model still identifies as itself (an AI assistant, an LLM, etc.), while refusing to answer the question.
    \item 1 if the response has no characteristics of the role and the model still identifies as itself (an AI assistant, an LLM, etc.), while attempting to answer the question.
    \item 2 if the model identifies as itself (an AI assistant, an LLM, etc.) but has some attributes of the role.
    \item 3 if the model is fully playing the role, either not mentioning being an AI or giving itself another name. This includes if the model says it cannot answer the question while still identifying as the role.
\end{itemize}

Respond with a number between 0 and 3. Don't say anything else, just the number.
\end{promptbox}
\caption{Evaluation prompt used for role-play quality scoring in the main pipeline. The judge model receives the target role, granularity level, user question, and generated answer, and assigns a score from 0 to 3 based on the degree of role adherence.}
\label{pmt:role-eval}
\end{figure*}

\begin{figure*}[ht]
\begin{promptbox}{Steering Evaluation Judge Prompt}
\textbf{Purpose.}
This evaluation prompt is used in the steering pipeline to assess how socially micro-versus-macro a generated response is after activation steering.
The judge model receives the original prompt and the steered model response, and rates the response along multiple granularity-related dimensions.

\medskip 

You are evaluating how socially micro-vs-macro a response is.

Rate the response on the following dimensions:
\begin{itemize}
    \item \texttt{"temporal\_scope"}: Temporal scope. Score 1--5 where 1 = ``immediate, day-to-day, short-term coping'' and 5 = ``long-term, strategic, intergenerational planning''.
    \item \texttt{"collectivity"}: Collectivity. Score 1--5 where 1 = ``individuals, families, or very small groups'' and 5 = ``organizations, institutions, populations, or international actors''.
    \item \texttt{"abstraction"}: Abstraction. Score 1--5 where 1 = ``concrete lived experience and specific local details'' and 5 = ``abstract policy, system design, and structural framing''.
    \item \texttt{"decision\_logic"}: Decision logic. Score 1--5 where 1 = ``personal coping, intuition, and ad hoc problem solving'' and 5 = ``formal planning, governance, incentives, and coordinated strategy''.
    \item \texttt{"first\_personness"}: First-personness. Score 1--5 where 1 = ``strongly first-person, self-referential, personal voice'' and 5 = ``impersonal, collective, institutional, or analytic voice''.
    \item \texttt{"uncertainty"}: Uncertainty. Score 1--5 where 1 = ``hesitant, tentative, or emotionally unsure'' and 5 = ``confident, declarative, and policy-like''.
    \item \texttt{"granularity\_overall"}: 1--5. Overall placement from micro to macro.
    \item \texttt{"degeneration"}: 0 or 1. Use 1 if the response is clearly repetitive, incoherent, or degraded.
\end{itemize}

\textbf{Prompt:}

[QUESTION START]
\{prompt\}
[QUESTION END]

\textbf{Model response:}

[ANSWER START]
\{response\}
[ANSWER END]

Return ONLY valid JSON with numeric fields for all requested ids.
Do not include markdown fences or extra commentary.
\end{promptbox}
\caption{Judge prompt used for steering evaluation. The judge model rates each steered response along six granularity-related dimensions, together with an overall granularity score and a degeneration flag, and returns only a JSON object.}
\label{pmt:steering-eval}
\end{figure*}

\end{document}